\documentclass[12pt]{article}
\usepackage{fullpage,graphicx,psfrag,amsmath,amsfonts,verbatim}
\usepackage[small,bf]{caption}\usepackage{amssymb}
\usepackage{enumitem}
\usepackage{tikz}
\usepackage{listings}
\usepackage{authblk}
\usepackage{subfiles}
\usepackage{url}
\usepackage{textcomp}
\usepackage{booktabs}
\usepackage{algorithm}
\usepackage{algpseudocode}
\usepackage{caption}
\usepackage{subfig}
\usepackage{hyperref}
\usepackage{gensymb}

\usepackage{xcolor}
\hypersetup{
    colorlinks=true,     
    linkcolor=blue,      
    citecolor=red,       
    urlcolor=magenta     
}

\newtheorem{lemma}{Lemma}
\newtheorem{proposition}{Proposition}

\lstdefinestyle{mystyle}{
      backgroundcolor=\color{backcolour},
      keywordstyle=\color{magenta},
      numberstyle=\tiny\color{codegray},
      stringstyle=\color{codepurple},
      basicstyle=\ttfamily\small,
      breakatwhitespace=false,
      breaklines=true,
      captionpos=t,
      keepspaces=true,
      numbers=left,
      numbersep=5pt,
      showspaces=false,
      showstringspaces=false,
      showtabs=false,
      tabsize=2,
}

\newcommand{\reals}{{\mbox{\bf R}}}
\newcommand{\complex}{{\mbox{\bf C}}}

\newcommand{\symm}{{\mbox{\bf S}}}  
\newcommand{\herm}{{\mbox{\bf H}}}

\newcommand{\Tr}{\mathop{\bf Tr}}

\newcommand{\diag}{\mathop{\bf diag}}

\newcommand{\Expect}{\mathop{\bf E{}}}
\newcommand{\Prob}{\mathop{\bf Prob}}

\newcommand{\argmax}{\mathop{\rm argmax}}

\newcommand{\ie}{{\it i.e.,}}

\newcommand{\BEAS}{\begin{eqnarray*}}
\newcommand{\EEAS}{\end{eqnarray*}}
\newcommand{\BEA}{\begin{eqnarray}}
\newcommand{\EEA}{\end{eqnarray}}
\newcommand{\BEQ}{\begin{equation}}
\newcommand{\EEQ}{\end{equation}}
\newcommand{\BIT}{\begin{itemize}}
\newcommand{\EIT}{\end{itemize}}
\newcommand{\BNUM}{\begin{enumerate}}
\newcommand{\ENUM}{\end{enumerate}}

\newcommand{\BEQN}{\begin{equation*}}
\newcommand{\EEQN}{\end{equation*}}

\newcounter{algorithmctr}
\renewcommand{\thealgorithmctr}{\arabic{algorithmctr}}
\newenvironment{algdesc}%
   {\mbox{}\\*[\parskip]\begin{minipage}{\linewidth}%
       \refstepcounter{algorithmctr}\begin{list}{}{%
       \setlength{\rightmargin}{0\linewidth}%
       \setlength{\leftmargin}{.05\linewidth}}%
       \rmfamily\small
       \item[]{\setlength{\parskip}{0ex}\hrulefill\par%
        \nopagebreak{\bfseries\textsf{Algorithm \thealgorithmctr~}}}}%
   {{\setlength{\parskip}{-1ex}\nopagebreak\par\hrulefill\\*[2ex]\par}%
   \end{list}\end{minipage}}

\begin{document}
\title{\texttt{T-Rex}: Fitting a Robust Factor Model via Expectation-Maximization}
\author[]{Daniel Cederberg}

\maketitle

\begin{abstract}
Over the past decades, there has been a surge of interest in studying
low-dimensional structures within high-dimensional data. Statistical factor
models --- \ie~low-rank plus diagonal covariance structures --- offer a
powerful framework for modeling such structures. However, traditional
methods for fitting statistical factor models, such as principal component
analysis (PCA) or maximum likelihood estimation assuming the data is
Gaussian, are highly sensitive to heavy tails and outliers in the observed
data. In this paper, we propose a novel expectation-maximization (EM)
algorithm for robustly fitting statistical factor models. Our approach is
based on Tyler's M-estimator of the scatter matrix for an elliptical
distribution, and consists of solving Tyler's maximum likelihood estimation
problem while imposing a structural constraint that enforces the low-rank
plus diagonal  covariance structure. We present numerical experiments on
both synthetic and real examples, demonstrating the robustness of our
method for direction-of-arrival estimation in nonuniform noise and subspace
recovery. 
\end{abstract}


\section{Introduction}
\label{sec:introduction}
Traditional statistical signal processing algorithms commonly assume that
the observed data follows a Gaussian distribution. While this model is
suitable for many applications, it tends to make those algorithms highly
sensitive to deviations from Gaussianity, resulting in severe performance
degradation when the observed data contains heavy tails or outliers
\cite{Zoubir2012}. The field of \emph{robust statistical signal processing}
\cite{Zoubir2018} seeks to overcome these limitations by developing tools
that remain effective even when the observed data deviates significantly
from the Gaussian distribution.

In this paper, we consider the problem of robustly estimating a
covariance matrix $\Sigma \in \symm^n$ of the form 
\BEQ \label{e:covariance-model}
\Sigma = FF^T + D,
\EEQ 
where $D \in \reals^{n \times n}$ is diagonal and $F \in \reals^{n \times r}$
with $r \leq n$. This covariance structure, known as a \emph{statistical factor
model}, is closely related to factor analysis \cite{basilevsky1994}, a
technique used to capture underlying relationships in high-dimensional data with
a parsimonious set of parameters. Statistical factor models have widespread
applications across various fields, including econometrics, finance, and signal
processing \cite{Bartholomew2011, Goodfellow2016, Fan2021}. There also exist
macroeconomic and fundamental factor models \cite{connor1995}, but in this paper
we will restrict our attention to fitting \emph{statistical} factor models.

There are two main techniques for fitting a statistical factor model. The first
technique starts by computing an unstructured covariance estimate $\hat{\Sigma}
\in \symm^n$, such as the sample covariance matrix, followed by a second step
that decomposes the preliminary estimate $\hat{\Sigma}$ according to
\eqref{e:covariance-model}. A famous method for the decomposition step is
based on principal components analysis (PCA) \cite[\S9.3]{johnson2007applied}, and is
closely related to least-squares approaches that fit a statistical factor model
by minimizing the Frobenius norm difference $\|\hat{\Sigma} - (FF^T +
D)\|^2_{F}$ with respect to $F$ and $D$ \cite{Harman1966, Joreskog1972,
Liao2016, Stoica2023}. These decomposition methods have in common that they make
no explicit assumption on the distribution of the observed data. However, if the
data contains heavy tails or outliers, using the sample covariance matrix as the
preliminary estimate leads to poor performance \cite{PISON2003}, since the
sample covariance matrix tends to be a poor estimate of the true covariance
matrix for heavy-tailed data.

The second technique for fitting a covariance matrix of the form
\eqref{e:covariance-model} assumes that the distribution of the observed data
is known. The matrices $F$ and $D$ are then estimated by enforcing the structure
\eqref{e:covariance-model} into the maximum likelihood (ML) estimation problem under
the assumed distribution. It is most common to model the data as Gaussian, 
resulting in the ML estimation problem \cite{Joreskog1967} 
\BEQ 
\label{e:Gaussian-FA-prob}
 \mbox{minimize }  \log \det (FF^T + D)  + \Tr(S
(FF^T + D)^{-1}) 
\EEQ 
with variables $F \in \reals^{n \times r}$ and diagonal $D \in \reals^{n \times
n}$. (The problem data is the sample covariance matrix $S \in \symm^n_+.)$ A
popular algorithm for solving \eqref{e:Gaussian-FA-prob} is the
expectation-maximization (EM) algorithm by Rubin \& Thayer \cite{Rubin1982},
which has a cheap iteration cost of order $\mathcal{O}(n^2 r)$. Alternative
methods for solving \eqref{e:Gaussian-FA-prob} have also been proposed
\cite{Liao2016, Stoica2023, liu1998,liu1998a,Zhao2008,Khamaru2019}. However, most
of these methods compute a spectral decomposition in each iteration, incurring a
higher iteration cost of order $\mathcal{O}(n^3)$ and thus limiting their
scalability compared to Rubin \& Thayer's EM. If the diagonal matrix $D$ is
constrained to have all entries equal, problem \eqref{e:Gaussian-FA-prob} has a
closed-form solution and is called \emph{probabilistic PCA} \cite{tipping1999}. 

A drawback of assuming 
an underlying Gaussian distribution is that the estimation of $F$ and $D$ tends to become
sensitive to outliers or heavy tails in the observed data. To accomodate for more 
heavy-tailed data, extensions based on the multivariate $t$-distribution 
\cite{Liu1995, Zhang2014, Zhou2019}, and other elliptical
distributions \cite{hippert2023learning, Hippert2021}, have
been proposed. Extensions of probabilistic
PCA to heteroscedastic data have also been studied 
\cite{hong2021heppcat, Collas2021, Zhang2022}.

In this paper, we introduce a new method for robustly fitting a statistical
factor model. The derivation of our method assumes that the distribution of the
random variable $Z$ describing the observed data belongs to the broad family of
\emph{elliptical probability distributions}. This family, which encompasses the
Gaussian distribution, preserves the elliptical symmetry of the Gaussian density
while allowing for distributions with much heavier tails. We then estimate $F$
and $D$ using a maximum likelihood approach based on the normalized random
variable $Z / \|Z\|_2$. Our method can be viewed as a variant of Tyler's
M-estimator for the scatter matrix of an elliptical distribution
\cite{Tyler1987}, adapted to the covariance structure
\eqref{e:covariance-model}. Tyler's estimator has garnered significant attention
in the signal processing and optimization communities \cite{Tyler1987, Chen2011,
Sun2014, Wiesel2015, Zhang2016, Frank2020, Goes2020, Ollila2021, Danon2022}. A
popular line of research has been to incorporate structural constraints into the
ML estimation problem associated with Tyler's estimator \cite{Soloveychik2014,
Sun2015, Soloveychik2016a, Bouchard2021}. The existing works \cite{Sun2015,
Bouchard2021} investigate Tyler’s ML estimation problem under the structural
constraint \eqref{e:covariance-model} with the additional assumption that the
diagonal entries of $D$ are equal.

While Tyler's estimator has been extensively studied, no work seems to have
investigated the impact of incorporating the covariance structure
\eqref{e:covariance-model} into the ML estimation problem associated with
Tyler's estimator. In this paper we close this gap by introducing \texttt{T-Rex}
(\textbf{T}yler's \textbf{R}obust Factor Model via
\textbf{E}xpectation-Ma\textbf{X}imization), a novel EM algorithm for robustly
fitting a statistical factor model. (As a surprising byproduct of our novel
EM-based analysis, we also uncover that Tyler's classical fixed-point iteration
is itself an instance of the EM algorithm.) We empirically compare
\texttt{T-Rex} against factor model estimators that assume the observed data
follows either a Gaussian or multivariate $t$-distribution. The numerical
results show that \texttt{T-Rex} is robust, matching the performance of
Gaussian-based methods on Gaussian data, and performing as well as methods based
on the multivariate $t$-distribution on multivariate $t$-distributed data.
Furthermore, our method outperforms the competing estimators when the data is
contaminated with outliers or deviates from the assumed data distributions of
the competing estimators. We also highlight the utility of \texttt{T-Rex} for
robust direction-of-arrival estimation \cite{Mecklenbrauker2024} in non-uniform
noise \cite{Liao2016}, and robust subspace recovery \cite{Lerman2018}.

A closely related work to ours is the recent \emph{subspace constrained Tyler's
estimator} (\texttt{STE}) \cite{yu2024subspace}, which, like \texttt{T-Rex},
integrates low-rank structure into Tyler's estimator. However, there are several
key differences between \texttt{T-Rex} and \texttt{STE}. First, \texttt{T-Rex}
has a clear interpretation as a maximum likelihood estimator, whereas
\texttt{STE} is derived heuristically. Specifically, at each iteration of
Tyler's standard fixed-point procedure, \texttt{STE} applies a low-rank
approximation of the current scatter matrix estimate by computing its spectral
decomposition. This approach lacks a principled interpretation as a maximum
likelihood estimator. Second, \texttt{T-Rex} is designed for robust low-rank
scatter matrix estimation, while \texttt{STE} primarily focuses on the related task
of robust subspace recovery. Third, \texttt{T-Rex} seems to be more scalable
than \texttt{STE} since, unlike \texttt{STE}, it does not require a spectral
decomposition in each iteration. (We comment more on this in \S\ref{sec:num exp
robust subspace recovery}.) Finally, and most importantly, \texttt{T-Rex}
outperforms \texttt{STE} in our experiments for both robust covariance
estimation and subspace recovery in direction-of-arrival estimation.

The rest of the paper is organized as follows. In \S\ref{sec:preliminaries} we present
background on EM and Tyler’s estimator. In \S\ref{sec:T-Rex} we introduce our
approach for robustly fitting a statistical factor model, and derive the corresponding
EM algorithm \texttt{T-Rex}. In \S\ref{sec:num exp} we present extensive 
numerical experiments, followed by conclusions in \S\ref{sec:conclusions}.

\section{Preliminaries}
The technical foundation of this paper relies on Tyler's estimator and the
EM algorithm. In this section, we provide
background on both.

\label{sec:preliminaries}

\subsection{Background on Tyler's estimator}
Several common probability distributions, such as the Gaussian and the
multivariate $t$-distribution, are \emph{elliptical}. The probability
density function (pdf) of an elliptical $n$-dimensional random variable $Z$
with zero mean is of the form 
\BEQ \label{e:density-elliptical}
p(z; \Sigma) = \frac{C}{\sqrt{\det \Sigma}} g(z^T \Sigma^{-1} z)
\EEQ 
where $C > 0$ is a normalization constant, $\Sigma \in \symm^{n}_{++}$ is
the \emph{scatter matrix} and $g:\reals \to \reals_+$ is the \emph{density
generator} determining the decay of the tails \cite[\S 4.2]{Zoubir2018}.
(The scatter matrix is a positive multiple of the covariance matrix
whenever the latter exists.) If $Z$ is elliptical with zero mean, then the
normalized random variable $X = Z / \|Z\|_2$ has a so-called \emph{angular
central Gaussian distribution}, with a pdf of the form 
\BEQ \label{e:density-acgd}
p(x; \Sigma) = \frac{\Gamma(n/2)}{2\pi^{n/2} \sqrt{\det \Sigma}}  (x^T \Sigma^{-1} x)^{-n/2},
\EEQ 
where $\Gamma(\cdot)$ is the gamma function \cite[\S 4.2]{Zoubir2018}. A
remarkable feature of the density $p(x; \Sigma)$ is that it is independent
of the original elliptical distribution, since it does not depend on the
density generator $g$. The ML estimate of the scatter matrix $\Sigma$ given
$m$ independent observations $x_1, \dots, x_m$ of $X$  can be found by
solving 
\BEQ \label{e:Tyler-vanilla-opt-prob}
\text{minimize } \log \det \Sigma + \frac{n}{m} \sum_{i=1}^m \log(x_i^T \Sigma^{-1} x_i)
\EEQ
with variable $\Sigma \in \symm^n$. This estimator of the scatter matrix is
called \emph{Tyler's M-estimator} \cite{Tyler1987} and has garnered significant
attention in the signal processing and optimization communities (see the
references in \S\ref{sec:introduction}).

\subsection{Background on expectation-maximization}
\label{sec:background-EM}
The EM algorithm \cite{dempster1977} is useful for solving certain ML
estimation problems where direct maximization of the likelihood is
difficult. Suppose we have $m$ independent observations $\mathcal{X}
\triangleq \{x_1, \dots, x_m\}$ of a random variable $X$, and our goal is
to estimate a parameter vector $\theta$ parameterizing its density
function $p(x; \theta)$. The ML estimate of $\theta$ is the maximizer of
the log-likelihood function 
\[
L(\theta; \mathcal{X}) \triangleq \sum_{i=1}^m \log p(x_i; \theta).
\]
The EM algorithm can be useful when it is difficult to directly maximize
the log-likelihood with respect to $\theta$. The idea is to augment the
\emph{observed} data $\mathcal{X}$ with \emph{unobserved} data $\mathcal{R}
\triangleq \{R_1, \dots, , R_m\}$, where we model $R_i, \: i = 1, \dots, m$
as random variables. We further assume that they are independent and that
their pdf $p(r; \theta)$ depends on the parameter $\theta$. The key is to
augment the observed data in such a way that the \emph{complete
log-likelihood function}
\BEQ \label{e:complete-log-likelihood-function}
L(\theta; \mathcal{X}, \mathcal{R}) \triangleq \sum_{i=1}^m \log p(x_i, R_i; \theta)
\EEQ
would be easy to maximize if $\mathcal{R}$ were known. However, since
$\mathcal{R}$ is not known in practice, $L(\theta; \mathcal{X},
\mathcal{R})$ is a random variable; thus, maximizing it with respect to
$\theta$ is not possible. 

Given a parameter estimate $\theta_k$, the EM algorithm approximates the
observed data likelihood $L(\mathcal{X}; \theta)$ by taking the expectation
of the complete log-likelihood $L(\theta; \mathcal{X}, \mathcal{R})$,
conditioned on the observed data $\mathcal{X}$, under the assumption that
$\theta_k$ is the true parameter governing the distributions. In other
words, EM is based on the approximation 
\[
\begin{split}
L(\theta; \mathcal{X}) & \approx \Expect_{\mathcal{R}| \mathcal{X}, \theta_k} 
[L(\theta; \mathcal{X}, \mathcal{R})] = \sum_{i=1}^m \Expect_{R_i| x_i, \theta_k} [\log p(x_i, R_i; \theta)]
\end{split}
\]
where 
\[
\Expect_{R_i| x_i, \theta_k} [\log p(x_i, R_i; \theta)] = \int \log p(x_i, r; \theta) p(r | x_i, \theta_k) dr.
\] 
The next parameter estimate $\theta_{k+1}$ is then obtained from 
\[
\theta_{k+1} = \argmax_{\theta} \Expect_{\mathcal{R}| \mathcal{X}, \theta_k} [L(\theta; \mathcal{X}, \mathcal{R})].
\]

\section{An EM algorithm for fitting a robust factor model}
\label{sec:T-Rex}
In this section we describe a new method for fitting a statistical factor
model based on Tyler's estimator. Our approach is based on incorporating
the structural constraint \eqref{e:covariance-model} into problem
\eqref{e:Tyler-vanilla-opt-prob}, resulting in the problem 
\BEQ \label{e:Tyler-prob-factor-model}
\begin{split}
\text{minimize } & f(F, D) \triangleq \log \det (FF^T + D) +  
 \frac{n}{m} \sum_{i=1}^m
\log(x_i^T(FF^T + D)^{-1} x_i)
\end{split}
\EEQ 
with variables $F \in \reals^{n \times r}$ and diagonal $D \in \reals^{n
\times n}$. Problem \eqref{e:Tyler-prob-factor-model} is the ML estimation
problem for fitting a statistical factor model under the angular central
Gaussian distribution \eqref{e:density-acgd}. To solve \eqref{e:Tyler-prob-factor-model} we derive
an EM algorithm in \S\ref{sec:data-model-EM} and
\S\ref{sec:cond-expectation-EM}, followed by a summary and implementation
details in \S\ref{sec:EM-summary} and \S\ref{sec:EM-implementation},
respectively. 

\subsection{The latent variable model}
\label{sec:data-model-EM}
When fitting a statistical factor model, the unknown parameter to be
estimated is $\theta \triangleq (F, D)$. To design an EM algorithm for
solving \eqref{e:Tyler-prob-factor-model}, we introduce a probabilistic
model for the observed and unobserved variables, $X$ and $R$, in a way that
ensures $X$ is marginally distributed according to the angular central
Gaussian distribution \eqref{e:density-acgd}, and also allows for easy
maximization of the complete data likelihood if $\mathcal{R}$ were known.
We specify the latent variable model by assuming that the joint pdf of $(X,
R)$ is 
\BEQ \label{e:latent-variable-model}
p(x, r; \theta) = \frac{r^{n-1}}{(2\pi)^{n/2} \sqrt{\det \Sigma_{FD}}} \exp 
\left(-\frac{r^2}{2} x^T \Sigma_{FD}^{-1} x  \right)
\EEQ 
where $\Sigma_{FD} \triangleq FF^T + D$, with support $\mathbb{S}^{n-1}
\times \reals_+$ where $\mathbb{S}^{n-1}$ is the unit sphere in $\reals^n$.
(We motivate this choice of joint pdf heuristically in Appendix
\ref{sec:latent-variable-model-appendix}; roughly speaking, it can be
interpreted as the pdf of a random variable $Z \sim N(0, \Sigma_{FD})$
parameterized as $Z = RX$ where $X$ has unit norm.) Given this
specification for the joint density, the marginal density of $X$ is 
\[
\begin{split}
 p(x; \theta)  & = \int_0^\infty p(x, r; \theta) dr  \\
& = \frac{1}{(2\pi)^{n/2} \sqrt{\det \Sigma_{FD}}} \int_0^\infty r^{n-1} \exp \left(-\frac{r^2}{2} x^T \Sigma_{FD}^{-1} x\right) dr \\
& = \frac{1}{(2\pi)^{n/2} \sqrt{\det \Sigma_{FD}}} \frac{\Gamma(n/2) 2^{n/2-1}}{(x^T \Sigma_{FD}^{-1} x)^{n/2}} \\
& = \frac{\Gamma(n/2)}{2\pi^{n/2} \sqrt{\det \Sigma_{FD}}} (x^T \Sigma_{FD}^{-1} x)^{-n/2},
\end{split}
\]
where we in the third equality have used that  
\cite[equation (3.326)]{gradshteyn2007}
\[
\int_0^\infty r^{n-1} \exp(-(a/2) r^2) dr = \frac{\Gamma(n/2) 2^{n/2-1}}{a^{n/2}}
\] 
for any $a > 0$. Thus, we see that under the latent variable model
\eqref{e:latent-variable-model}, the marginal density of $X$ is the pdf of
the angular central Gaussian distribution \eqref{e:density-acgd}, which
means that an EM algorithm based on this latent variable model will indeed target
the ML estimation problem \eqref{e:Tyler-prob-factor-model}.
 
\subsection{The expectation and maximization steps}
\label{sec:cond-expectation-EM}
Under the latent variable model \eqref{e:latent-variable-model}, the
complete log-likelihood function \eqref{e:complete-log-likelihood-function}
is 
\BEQ \label{e:complete-data-likelihood}
\begin{split}
 L(\theta; \mathcal{X}, \mathcal{R}) & = m(n-1) \log r - (mn/2) \log(2\pi) 
- \frac{m}{2} \log \det \Sigma_{FD}- \frac{1}{2} \sum_{i=1}^m r_i^2 x_i^T \Sigma_{FD}^{-1} x_i \\
& = c - \frac{m}{2}\left( \log \det \Sigma_{FD} + \Tr (\Sigma_{FD}^{-1} S_{rx} )\right)
\end{split}
\EEQ
where $c \triangleq m(n-1) \log r - (mn/2) \log(2 \pi)$ is independent of
$(F, D)$, and 
\[
S_{rx} \triangleq (1/m)\sum_{i=1}^m r_i^2 x_i x_i^T.
\]
At the $k$th iteration of EM, we start with the current iterate $(F_k,
D_k)$. To compute the next iterate $(F_{k+1}, D_{k+1})$, we maximize the
complete log-likelihood \eqref{e:complete-data-likelihood}, replacing
the unobserved $S_{rx}$ with its conditional expectation given the observed
data $\mathcal{X}$. Specifically, $(F_{k+1}, D_{k+1})$ is the solution of 
\BEQ \label{e:subproblem}
\text{minimize } \log \det (FF^T + D) + \Tr((FF^T + D)^{-1} \hat{S}^{(k)})
\EEQ 
with variables $F \in \reals^{n \times r}$ and diagonal $D \in \reals^{n \times n}$, where 
\BEQ \label{e:expected-suff-stat}
\begin{split}
\hat{S}^{(k)} & \triangleq \Expect_{\mathcal{R}| \mathcal{X}, \theta_k}[S_{rx}]  =
\frac{1}{m}\sum_{i=1}^m \Expect_{R_i | x_i; F_k, D_k} [ r_i^2 x_i x_i^T] 
 = \frac{1}{m} \sum_{i=1}^m \Expect_{R_i | x_i; F_k, D_k} [r_i^2] x_i x_i^T.
\end{split}
\EEQ 
To evaluate the conditional expectation in \eqref{e:expected-suff-stat},
the following result showing that $R_i | x_i$ has the distribution of a
scaled chi-distributed random variable will be useful.

\begin{lemma} \label{lem:chi-distribution}
Assume the joint density of $(X, R)$ is given by
\eqref{e:latent-variable-model}. Then the conditional distribution of $R$
given $X = x$ is the same as the distribution of $(1/\sqrt{x^T
\Sigma_{FD}^{-1} x}) S$ where $S \sim \chi(n)$.
\end{lemma}
\emph{Proof.}
The conditional distribution of $R$ given $X = x$ is 
\[
\begin{split}
p(r | x; \theta) & = \frac{p(x, r; \theta)}{p(x; \theta)} \\
& =
\frac{\frac{1}{(2\pi)^{n/2} \sqrt{\det \Sigma_{FD}}} \exp\left(-\frac{r^2}{2} x^T \Sigma_{FD}^{-1} x\right) r^{n-1}}{\frac{\Gamma(n/2)}{2\pi^{n/2} \sqrt{\det \Sigma_{FD}}} (x^T \Sigma_{FD}^{-1} x)^{-n/2}} \\
& = \frac{(x^T \Sigma_{FD}^{-1} x)^{n/2}}{\Gamma(n/2) 2^{n/2-1}} \exp\left(-\frac{r^2}{2} x^T \Sigma_{FD}^{-1} x\right) r^{n-1}. \\
\end{split}
\]
If $S \sim \chi(n)$, then the pdf of $S$ is \cite[\S 8.14]{christian2007}
\[
p_S(s) = \frac{1}{\Gamma(n/2) 2^{n/2 - 1}} s^{n-1} \exp\left(-\frac{s^2}{2}\right).
\]
From the fact that the pdf of $aS$ with $a = 1/\sqrt{x^T \Sigma_{FD}^{-1} x}$ is $(1/a)p_S(s/a)$, we get the desired result.
$\blacksquare$

By using Lemma \ref{lem:chi-distribution} together with the fact that the
second moment of a $\chi(n)$-distributed random variable is $n$ 
\cite[\S 8.14]{christian2007}, we conclude that $\hat{S}^{(k)}$ in
\eqref{e:expected-suff-stat} can be computed as
\BEQ \label{e:main-result-cond-expect}
\hat{S}^{(k)} = \frac{n}{m} \sum_{i=1}^m \frac{1}{x_i^T (F_k F_k^T + D_k)^{-1} x_i} x_i x_i^T.
\EEQ

\subsection{The complete algorithm}
\label{sec:EM-summary}
We can now summarize the EM algorithm for solving
\eqref{e:Tyler-prob-factor-model}. The E-step consists of computing
$\hat{S}^{(k)}$ using \eqref{e:main-result-cond-expect}, and the M-step is
given by \eqref{e:subproblem}. The complete method for solving
\eqref{e:Tyler-prob-factor-model} is summarized in Algorithm
\ref{alg-Tyler-FA-EM} below. We call it \texttt{T-Rex}, standing for
\textbf{T}yler's \textbf{R}obust Factor Model via
\textbf{E}xpectation-ma\textbf{X}imization. We will later discuss the
initialization, how to evaluate the M-step, and the termination criterion.

\begin{algdesc}{ \sc T-Rex  }\label{alg-Tyler-FA-EM}
{\footnotesize
\begin{tabbing}
\emph{Initialization.} Compute $F_0$ and $D_0$ \\*[\smallskipamount]
{\bf repeat for $k = 0, 1, \dots$} \\
\qquad \= 1.\ \emph{E-step.} Compute $\hat{S}^{(k)}$ using \eqref{e:main-result-cond-expect}.  \\
\> 2.\ \emph{M-step.} Let $(F_{k+1}, D_{k+1})$ be the solution of
\eqref{e:subproblem}. \\
\> 3.\ Check termination criterion. 
\end{tabbing}}  
\end{algdesc}

\paragraph{Interpretation of \texttt{T-Rex}.} \texttt{T-Rex} has an
insightful interpretation. The M-step can be interpreted as fitting a
low-rank plus diagonal approximation of the matrix $\hat{S}^{(k)}$
\cite{Johansson23}. This matrix resembles an adaptively weighted sample
covariance matrix, where the contributions of rank-one terms $x_i x_i^T$
are downweighted when the quantity  $x_i^T (F_k F_k^T + D_k)^{-1} x_i$ is
large. The quantity $x_i^T (F_k F_k^T + D_k)^{-1} x_i$ is the
\emph{Mahalanobis distance} between $x_i$ and the origin with respect to
$F_k F_k^T + D_k$. In robust statistics, a large Mahalanobis distance is
typically interpreted as an indicator of an outlier. (This is easy to
justify for an elliptical distribution, since the density function
\eqref{e:density-elliptical} for an elliptical distribution takes on a
small value when the Mahalanobis distance is large.) Hence, roughly
speaking, the EM algorithm \emph{iteratively fits statistical factor models to weighted
sample covariance matrices, where outliers with respect to the current fit
have been downweighted}. 

\paragraph{Interpretation of Tyler's fixed-point iteration as EM.}
Tyler's estimator of a scatter matrix (without taking the factor model into
account) is traditionally defined as a positive definite matrix
$\Sigma^\star \in \symm^n$ satisfying \cite{Tyler1987}
\[
\Sigma^\star = \frac{n}{m} \sum_{i=1}^m \frac{1}{x_i^T (\Sigma^\star)^{-1} x_i} x_i x_i^T.
\]
(This equation is equivalent to setting the gradient of the objective
function of \eqref{e:Tyler-vanilla-opt-prob} to zero.) The traditional way
of computing Tyler's estimator is through the fixed-point iteration 
\BEQ \label{e:Tyler-FP}
\Sigma_{k+1} = \frac{n}{m}\sum_{i=1}^m \frac{1}{x_i^T \Sigma_k^{-1} x_i} x_i x_i^T.
\EEQ
If we derive an EM algorithm similar to \texttt{T-Rex} to compute Tyler's
estimator by solving \eqref{e:Tyler-vanilla-opt-prob}, the M-step is to
minimize the function $\Sigma \mapsto  \log \det \Sigma + \Tr(\Sigma^{-1} \hat{S}^{k})$
where $\hat{S}^{(k)} = (1/m)\sum_{i=1}^m (1/(x_i^T \Sigma_k^{-1} x_k)) x_i
x_i^T.$ This M-step is solvable in closed form and gives the EM update
$\Sigma_{k+1} = \hat{S}^{(k)}$, which is identical to the fixed-point
update \eqref{e:Tyler-FP}. This reveals that \emph{Tyler's renowned
fixed-point iteration can be interpreted as an EM algorithm.} Since this
interpretation seems to be new in the literature, we state it as a formal
result. 
\begin{proposition} \label{thm:Tyler-EM}
Consider the ML estimation problem \eqref{e:Tyler-vanilla-opt-prob} for
fitting an unstructured scatter matrix $\Sigma \in \symm^n$ for a random
variable $X$ following the angular central Gaussian distribution. The EM
algorithm obtained from the latent variable model 
\[
p(x, r; \theta) = \frac{r^{n-1}}{(2\pi)^{n/2} \sqrt{\det \Sigma}} \exp 
\left(-\frac{r^2}{2} x^T \Sigma^{-1} x  \right)
\]
where $R$ is the latent variable, is equivalent to Tyler's fixed-point
iteration \eqref{e:Tyler-FP}.
\end{proposition}
This interpretation is useful for developing faster methods to compute
Tyler's estimator, by incorporating acceleration techniques or general
improvements to EM algorithms \cite{jamshidian1997, liu1998}. It also
provides a framework for analyzing Tyler's fixed-point iteration and its
extensions \cite{Chen2011, Ollila2021} through the lens of EM.

\subsection{Implementation}
\label{sec:EM-implementation} 
\paragraph{Initialization and termination.}
To initialize \texttt{T-Rex} we use PCA, and a possible termination
criterion is based on the relative change in objective value between
consecutive iterates. More details are presented in Appendix
\ref{sec:Implementation details}.

\paragraph{Evaluating the M-step.}
The M-step of \texttt{T-Rex} is similar to problem
\eqref{e:Gaussian-FA-prob}, which is the ML estimation problem when the
observed data is assumed to be Gaussian. This problem has been extensively
studied (see the prior work in \S \ref{sec:introduction}) and can be solved
with several algorithms. We prefer the EM algorithm by Rubin \& Thayer
\cite{Rubin1982}, simply because it has iteration cost $\mathcal{O}(n^2 r)$
rather than $\mathcal{O}(n^3)$ and seems to work well in practice. We
summarize it in Appendix \ref{sec:Implementation details}.

\paragraph{Exploiting low-rank structure.} 
Several steps of \texttt{T-Rex} can be sped up using the inherent low-rank
structure. For example, the matrix inversion formula \cite[\S C.4]{boyd2004} can be
used to compute $\Sigma_k^{-1} x_i$ for $i = 1, \dots, m$, where $\Sigma_k
= F_k F_k^T + D_k$, at a cost of order $\mathcal{O}(nmr)$. This is
significantly lower than the cost when the low-rank structure is not taken
into account which is of order $\mathcal{O}(n^3) + \mathcal{O}(n^2m)$. 

Computing $\Sigma_k^{-1} x_i$ for $i = 1, \dots, m$ is the bottleneck in
the fixed-point iteration \eqref{e:Tyler-FP} used to compute Tyler's
estimator without the factor model structure. This bottleneck is the key
reason why computing the standard Tyler's estimator is often considered
computationally prohibitive in high-dimensional settings. However,
\texttt{T-Rex} exploits the low-rank structure and allows Tyler's estimator
to be scaled up more gracefully to higher dimensions.
 
\paragraph{Iteration cost.} 
In every iteration of \texttt{T-Rex} we compute $\Sigma_k^{-1} x_i, \: i =
1, \dots, m$ at a cost of order $\mathcal{O}(nmr)$, we then form
$\hat{S}^{(k)}$ at a cost of order $\mathcal{O}(n^2 m)$, and finally
evaluate the M-step using the EM algorithm from \cite{Rubin1982} with
iteration cost $\mathcal{O}(n^2 r)$. We have observed that \texttt{T-Rex}
in practice typically converges quickly to a local minimizer of
\eqref{e:Tyler-prob-factor-model}, often within 5-10 iterations.

In the application of subspace recovery studied in the next section, we
have $n \gg m$ with $n = 32256$ and $m = 499$. In this setting it is
possible to implement $\texttt{T-Rex}$ without forming any $n \times n$
matrices, including $\hat{S}^{(k)}$, explicitly. Furthermore, in this case
we can implement the EM algorithm by Rubin \& Thayer with an iteration cost
of order $\mathcal{O}(nmr)$, representing a substantial improvement over
the cost $\mathcal{O}(n^2r)$.

\section{Numerical experiments}
\label{sec:num exp}
In this section we evaluate the empirical performance of \texttt{T-Rex}. In the
first experiment, we test its performance on synthetic data generated from
various distributions. In the second experiment, we compare \texttt{T-Rex} to
the standard Tyler's estimator. In the third experiment, we demonstrate the
utility of \texttt{T-Rex} for robust direction-of-arrival estimation in
nonuniform noise. Finally, we apply \texttt{T-Rex} to a stylized application of
robust subspace recovery. The code is available at
\url{https://github.com/dance858/T-Rex-EM}.
 
\subsection{Synthetic data simulations}
\label{sec:num exp synthetic}

\paragraph{Experimental setup.}
First we consider an example with
synthetic data, where we assume that the true covariance matrix  
$\Sigma_{\text{true}} \in \symm^n$ is known.
We compare \texttt{T-Rex} against:
\BNUM 
\item \emph{Gaussian factor analysis} (\texttt{GFA}): We solve the Gaussian ML
estimation
problem \eqref{e:Gaussian-FA-prob} using the EM algorithm in \cite{Rubin1982}.
\item \emph{Multivariate-t factor analysis} (\texttt{TFA}):  We use the
Expectation/Conditional Maximization Either (ECME) algorithm in \cite{Liu1995}
to solve the factor model fitting ML estimation problem assuming the data
follows a multivariate $t$-distribution.
\item \emph{Subspace constrained Tyler's estimator} (\texttt{STE}):
We apply the recent method from \cite{yu2024subspace} that, like \texttt{T-Rex},
integrates low-rank structure into Tyler's estimator. Note, however, that
\texttt{STE} is not specifically designed for covariance estimation but rather
for the related task of robust subspace recovery.
\ENUM 
To demonstrate the performance of the estimators and highlight their
distinguishing features, we consider three scenarios. First, we consider
\emph{Gaussian data} where we draw observations of $x \sim N(0,
\Sigma_{\text{true}})$. Secondly, we consider \emph{heavy-tailed data} where we
draw observations from a zero mean multivariate $t$-distribution with covariance
matrix $\Sigma_{\text{true}}$ and three degrees of freedom. Lastly, we consider
\emph{contaminated Gaussian data} where we introduce a few outliers into the
Gaussian dataset from the first scenario. These outliers are drawn from $N(\mu,
\Sigma_{\text{true}})$, where each entry of the outlier mean vector $\mu$ has
magnitude $(3/\sqrt{n}) \sqrt{\Tr(\Sigma_{\text{true}})}$. (In the univariate
case when $n = 1$, this corresponds to drawing outliers with a mean equal to 3
standard deviations.)

To construct the true covariance matrix we apply PCA to financial data from S\&P
500 (more details are given in Appendix
\ref{sec:exp-details-covariance-construction}). We use $r = 5$ factors and
dimension $n = 50$, and vary the number of samples $m$. For the contaminated
data we add 2\% outliers. 
We measure the performance in terms of the average estimation error of the
correlation matrix. Specifically, we define the mean-squared-error (MSE) as \BEQ
\label{e:MSE-synthetic} \text{MSE} = \Expect \bigg[ \frac{\|
\hat{\Sigma}_{\text{corr}} - \Sigma_{\text{corr}} \|_F}{\| {\Sigma}_\text{corr}
\|_F}  \bigg], \EEQ where $\| \cdot \|_F$ is the Frobenius norm,
$\Sigma_\text{corr}$ is the true correlation matrix, and
$\hat{\Sigma}_\text{corr}$ is an estimate of the correlation matrix obtained by
normalizing the covariance or scatter matrix estimate obtained from either
\texttt{GFA}, \texttt{TFA}, \texttt{T-Rex}, or \texttt{STE}. The expectation in
\eqref{e:MSE-synthetic} is evaluated by averaging the errors over 100
independent realizations of the experiment.

\begin{figure}[htb]
\centering
\subfloat{\includegraphics[width=0.85\textwidth]{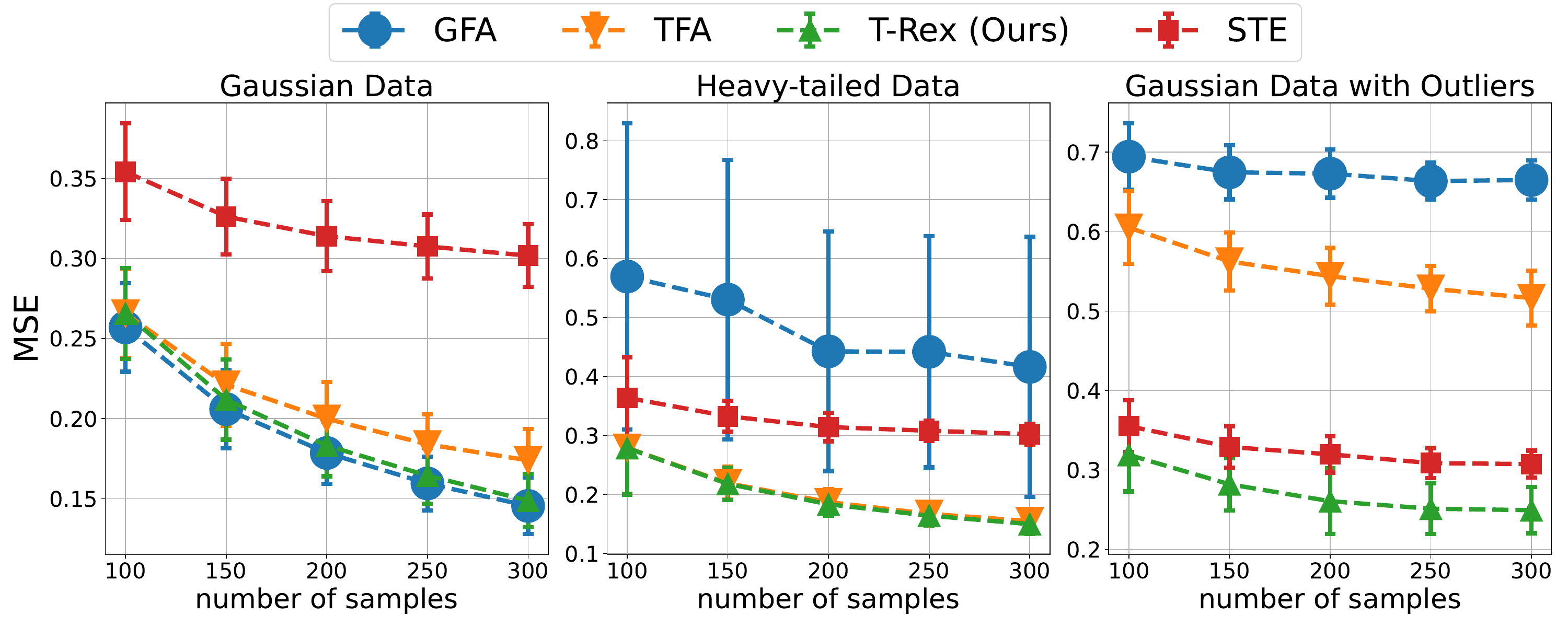}}
\caption{The MSE \eqref{e:MSE-synthetic} on three different types of data.
The error bars represent one standard deviation above and below the average
error over 100 runs.}
\label{fig:synthetic-data-MSE}
\end{figure}

\paragraph{Results.}
The result is presented in Figure
\ref{fig:synthetic-data-MSE}. We see that for Gaussian data, \texttt{GFA} achieves the lowest average error closely followed by \texttt{T-Rex}. For $m = 300$, \texttt{TFA}, \texttt{T-Rex} and \texttt{STE} achieve errors 
that are 19.6\%, 2.3\%, and 107\% larger than that of \texttt{GFA}, 
respectively. However, the performance of \texttt{GFA} deteriorates 
significantly on heavy-tailed data, with the MSE increasing with a factor of three. In contrast, \texttt{TFA}, \texttt{T-Rex} and \texttt{STE} 
remain unaffected. On Gaussian data with outliers, both \texttt{GFA} and 
\texttt{TFA} show a significant performance degradation, while 
\texttt{T-Rex} maintains its performance. This result demonstrates 
excellent empirical performance and highly robust properties of
\texttt{T-Rex}, as it performs reliably across different data.

\subsection{Comparison to Tyler's estimator}
In this experiment we compare \texttt{T-Rex} and the standard Tyler's estimator
\eqref{e:Tyler-FP} (\texttt{Tyler}) in terms of runtime and estimation accuracy.
As alluded to in \S\ref{sec:EM-implementation}, \texttt{T-Rex} has the potential
to speed up Tyler's estimator, allowing a Tyler-like estimator to be applied to
high-dimensional data. We confirm this speedup empirically, and also demonstrate
that incorporating a factor model structure can substantially improve the
quality of the estimated scatter matrix.

\begin{figure}[htb]
\centering
\subfloat{\includegraphics[width=0.75\textwidth]{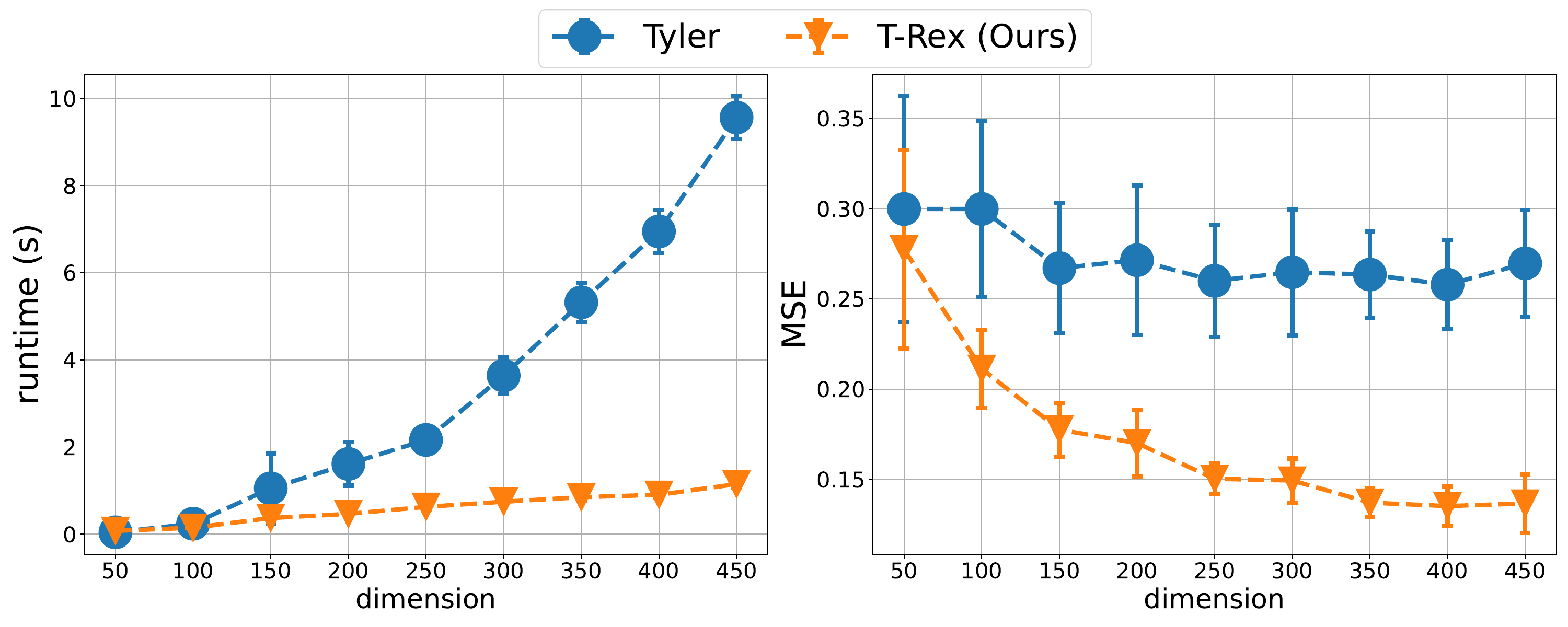}} 
\caption{\emph{Left.} The runtime of \texttt{T-Rex} and \texttt{Tyler} versus the dimension $n$ of the covariance. We ran both methods for a fixed number of 20 iterations. \emph{Right.} The MSE of \texttt{T-Rex} and \texttt{Tyler}.}
\label{fig:comp-Tyler-Trex}
\end{figure}

We consider Gaussian data where we draw observations of $x \sim N(0,
\Sigma_{\text{true}})$. The true covariance matrix is chosen to be the sample
covariance matrix of daily returns of $n$ stocks on S\&P 500 over the period
January 1, 2022 to January 1, 2024. In other words, we do not plant a low-rank
plus diagonal covariance structure in this experiment. For each dimension $n$,
we consider a challenging scenario with $m = n + 1$, \ie~the number of
observations is essentially the same as the dimension. We evaluate the MSE
\eqref{e:MSE-synthetic} as in the previous section. In Figure
\ref{fig:comp-Tyler-Trex} we see that \texttt{T-Rex} is occasionally nearly
twice as accurate as \texttt{Tyler}, and significantly faster in higher
dimensions. Moreover, unlike Tyler's standard estimator, the primary
computational cost of \texttt{T-Rex} is matrix-matrix multiplications --- making
it well-suited for GPU acceleration when runtime is a critical factor. We ran
\texttt{T-Rex} with $r = 5$ factors, and noted that the results were very
similar for $r = 10$. 

\subsection{Direction-of-arrival estimation in nonuniform noise}
In our next example we consider the problem of estimating the arrival angles of $r$ sources
emitting signals that impinge on a uniform linear array of $n$ sensors. The 
arrival angle of source $k$ relative the array normal direction is $\theta_k$,
and we model the array output $x_t \in \complex^n$ as 
\[
\qquad \qquad \qquad \qquad x_t = A(\theta) z_t + \epsilon_t, \qquad t = 1, \dots, m
\] where $\theta \in \reals^r$ is the vector of arrival angles, $A(\theta) \in \complex^{n \times r}$
is a steering matrix that depends on the geometry of the array, $z_t \in
\complex^r$ is the vector of source signals, and $\epsilon_t \in \complex^{n}$
represents noise. Under the assumption that the source signal is uncorrelated
with the noise, the covariance matrix $\Sigma \in \herm^{n}$ of the array output is 
\BEQ \label{e:DOA-covariance}
\Sigma = A(\theta) P A(\theta)^H + Q,
\EEQ where $P \in \herm^n$ is the signal covariance matrix, $\sigma_s^2 > 0$
is the signal power, and $Q \in \herm^{n}$ is the noise covariance matrix that
is further assumed to be diagonal. (Here
$\herm^n$ is the space of Hermitian matrices of dimension $n$ and the superscript $H$
denotes the conjugate transpose.)
Traditional direction-of-arrival (DOA) estimators are derived under the assumption of uniform white noise, 
\ie~$Q = \sigma^2 I$ for some $\sigma^2 > 0$. This assumption implies that the
signal and noise subspaces can be correctly determined through the
spectral decomposition of the array covariance matrix $\Sigma$. However, this is no
longer true when the noise is nonuniform, \ie~when $Q$ is an arbitrary diagonal
matrix, since in this case the eigenvectors of $\Sigma$
associated with the $r$ largest eigenvalues do no longer span the signal
subspace. 

Designing DOA estimators for nonuniform noise has garnered significant
attention (see, for example, \cite{Pesavento2001, Liao2016} and the references therein).
One approach is to decompose an estimate $\hat{\Sigma} \in
\herm^n$ of the array covariance matrix according to 
\[
\hat{\Sigma} = FF^H + D,
\] and then apply traditional DOA estimators such as root-MUSIC \cite{stoica2005spectral} to the matrix
$FF^H$. This method is justified by observing that if $\hat{\Sigma}$ is equal to
the true array covariance matrix, then the
column space of $F = A(\theta) P^{1/2}$ is identical to the column space of the
steering matrix $A(\theta)$, which is equal to the signal subspace.

\paragraph{Benchmark methods.}
We compare the performance of 
the sample covariance matrix $(\texttt{SC})$ with \texttt{GFA},
\texttt{T-Rex} and \texttt{STE} for DOA-estimation in nonuniform noise. 
We apply root-MUSIC 
directly to the sample covariance matrix and the scatter matrix estimate 
obtained from \texttt{STE}, whereas for \texttt{GFA} and
\texttt{T-Rex}, we first estimate the matrices $F$ and $D$ from the available
snapshots, and then apply root-MUSIC to the matrix $FF^H$. As performance metric we
use the mean-squared error 
\BEQ \label{e:MSE-def}
\text{MSE} = \Expect \| \hat{\theta} - \theta \|_2^2
\EEQ where $\theta \in \reals^r$ is the vector of true arrival angles, and $\hat{\theta} \in \reals^r$
is the estimate obtained from MUSIC. 

In this application, the data $x_t$ is complex-valued. Although the density
function for complex-valued elliptical distributions differs slightly from that
of real-valued elliptical distributions (see \cite[\S4]{Zoubir2018}), the
estimation problem for Tyler’s estimator under the structural constraint $\Sigma
= FF^H + D$ remains unchanged. Furthermore, careful bookkeeping shows that our
approach \texttt{T-Rex} extends to the complex-valued setting with essentially
no changes. 

We omit \texttt{TFA} from these experiments because the implementation we use
(the R-package \texttt{fitHeavyTail} \cite{Palomar23}) does not support
complex-valued data.

The purpose of this experiment is not to conduct an extensive benchmarking of
state-of-the-art methods for DOA estimation in nonuniform noise. Instead, we
simply want to highlight the differences between our approach, one based on the
Gaussian distribution (\texttt{GFA}), and another based on integrating low-rank
structure into Tyler's estimator (\texttt{STE}). However, it is worth noting
that the \texttt{GFA} method we compare against is based on solving the same ML
estimation problem as the one considered in \cite{Liao2016, Stoica2023}, which
has been described as state-of-the-art for DOA estimation in non-uniform noise.

\paragraph{Experimental setup.}
We consider a scenario with $n = 15$ sensors, $r = 4$ sources, true arrival
angles $\theta = (0\degree, \: 5\degree, \: 10\degree, \:15\degree)$, and source
covariance matrix $P = \sigma^2_s I$ where $\sigma^2_s = 1$ is the signal
power. We will
first assume that the array output follows a complex-circular Gaussian
distribution $x \sim \complex N(0, \Sigma)$ where $\Sigma$ is given by 
\eqref{e:DOA-covariance}. This is the most common assumption for DOA estimation
and corresponds to the case where the source signal and the noise are assumed
to be independent and Gaussian. We also plot the trace of the 
Crámer-Rao bound (CRB) matrix that is derived in closed-form for the Gaussian setting with nonuniform 
noise in \cite[equation (36)]{Pesavento2001}.

To improve robustness of DOA--estimators, researchers have proposed to consider a more heavy-tailed signal or noise model
\cite{Mahot2013, Zoubir2018}.
Following \cite{Ollila2003, Mecklenbrauker2022, Mecklenbrauker2024}, we will
therefore also consider the case when the array output $x$ follows a zero-mean complex
multivariate $t$-distribution with three degrees of freedom and covariance
matrix $\Sigma$ given by \eqref{e:DOA-covariance}. In this case we evaluate the CRB using 
\cite[equation (18)]{Greco2013}.

The signal-to-noise ratio (SNR) is defined as \cite{Liao2016}
\[
\text{SNR} = \frac{\sigma_s^2}{n} \sum_{i=1}^n \frac{1}{\sigma_i^2}
\] where $\sigma_i^2 > 0$ is the noise variance for the $i$th sensor. Following \cite{Stoica2023}, we draw the 
noise variances independently from a uniform distribution on $(0,1)$ and then
rescale them uniformly to obtain SNR = 5 dB.
To approximate the mean-squared
error \eqref{e:MSE-def} we use 300 Monte Carlo runs.

\begin{figure}[!htb]
\centering
\includegraphics[width=0.80\textwidth]{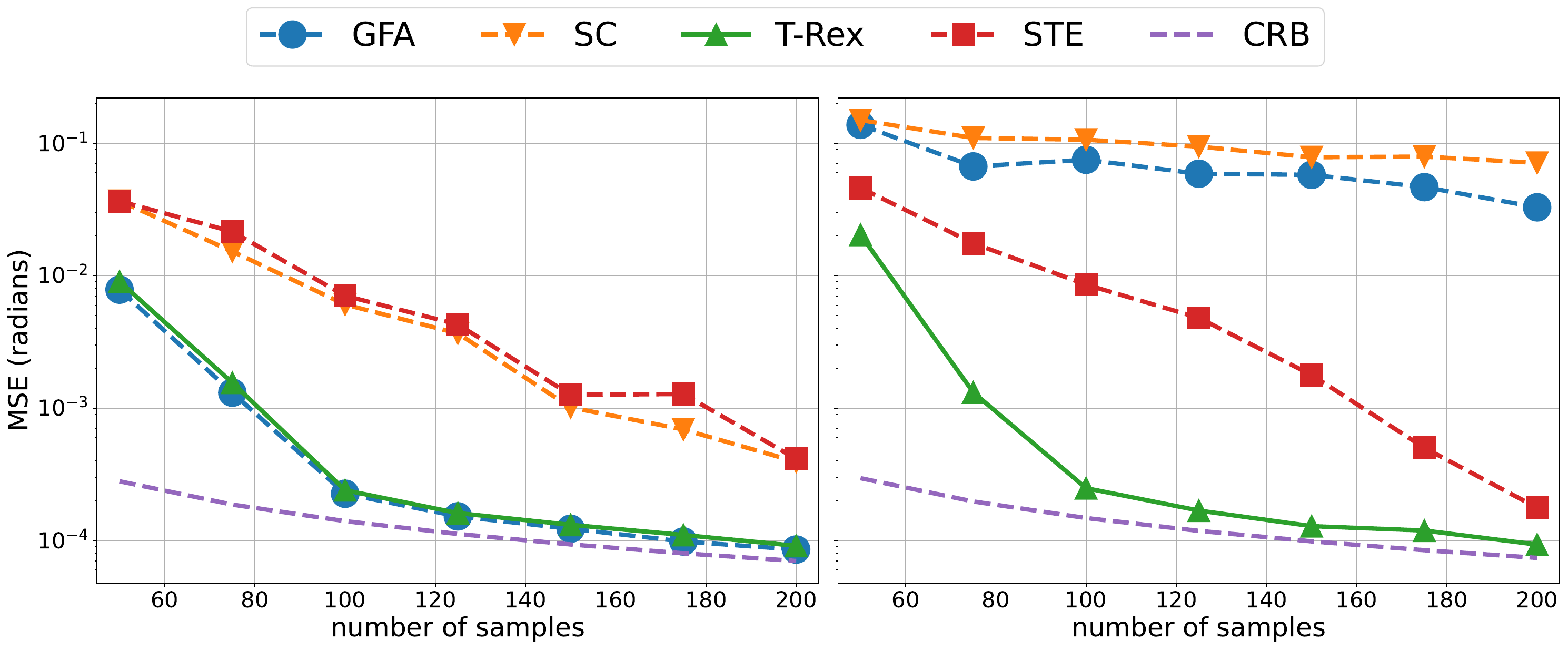}
\caption{\emph{Left.} The MSE versus the number of snapshots in the
Gaussian setting. \emph{Right.} The MSE versus the number of snapshots in
the heavy-tailed setting.}
\label{fig:DOA}
\end{figure}

\paragraph{Numerical results.}
Figure \ref{fig:DOA} presents the MSE versus the number of
snapshots for the Gaussian and multivariate $t$-distributed scenarios. In 
the Gaussian case, MUSIC combined with
the factor models obtained from \texttt{GFA} and \texttt{T-Rex} outperform
MUSIC applied directly to the sample covariance matrix. This is expected since
the noise is nonuniform. Notably, \texttt{T-Rex} and \texttt{GFA} exhibit
very similar performance in this setting. Furthermore, compared to 
\texttt{T-Rex} and \texttt{GFA}, the performance of \texttt{STE} is significantly worse.

\begin{figure}[!htb]
\centering
\includegraphics[width=0.70\textwidth]{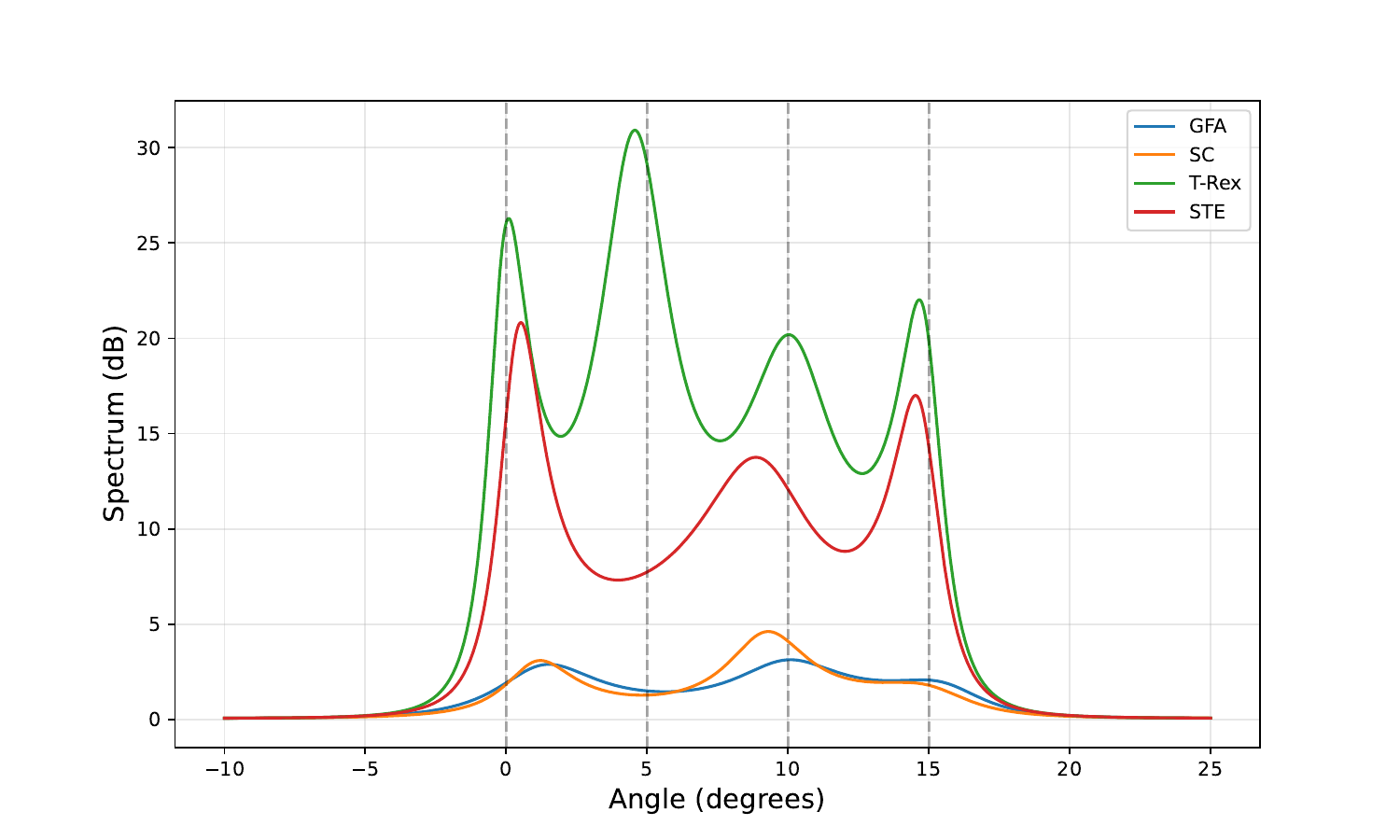}
\caption{Pseudospectrum for \texttt{T-Rex}, \texttt{G-FA}, \texttt{SC} and 
\texttt{STE} for $m = 100$ snapshots in the heavy-tailed scenario.}
\label{fig:pseudospectrum}
\end{figure}

In the heavy-tailed scenario in the right part of Figure \ref{fig:DOA}, however, the performance
of \texttt{GFA} deteriorates substantially, and it fails to resolve the
sources. In contrast, the robust approach \texttt{T-Rex} remains effective,
providing reliable estimates of the arrival angles. This is further
demonstrated in Figure \ref{fig:pseudospectrum}, where the pseudospectrum (see \cite[page 161]{stoica2005spectral}) is 
plotted for $m = 100$ snapshots. The pseudospectrum for \texttt{T-Rex}
clearly shows four peaks near the true arrival angles, whereas
\texttt{GFA}, \texttt{SC} and \texttt{STE} fail to identify four distinct peaks.

\subsection{Robust subspace recovery}
\label{sec:num exp robust subspace recovery}
In our final example we consider the problem of \emph{robust subspace recovery}
\cite{Lerman2018}. In particular, we demonstrate the ability of \texttt{T-Rex}
to robustly fit a low-dimensional linear subspace by using it to remove shadows
in images of faces.

\paragraph{Experimental setup.}
We consider the same setup as in
\cite{Lerman2015, Zhang2015} based on 64 images of a face from the extended Yale face
database \cite{Lee2005}. We use the first 32 images as the training set and
reserve the remaining 32 for the out-of-sample test. To learn a low-dimensional
subspace we use a dataset consisting of the 32 train images dispersed in a
collection of 467 random images from the Background/Google folder of the Caltech 
101 database \cite{FeiFei2004}. The 
images are 192 $\times$ 168 pixels.
By stacking the images as columns of
a matrix, we end up with a data matrix $X = \begin{bmatrix} x_1 & x_2 & \dots &
x_m \end{bmatrix}  \in \reals^{n \times m}$ with $m = 499$ samples, each sample
having dimension $n = 32256$. As is common when fitting a linear subspace
in a robust manner, we center the images by subtracting the so-called
\emph{Euclidean median}, which is defined as the minimizer of
$\sum_{i=1}^m \| x_i - c \|_2$ with respect to $c
\in \reals^n$.

Under the so-called \emph{Lambertian model}, images of convex objects taken
under changing illumination lie near a linear subspace of dimension $r = 9$
\cite{Basri2003}. (The Lambertian model is widely used in computer vision. It
describes how certain surfaces reflect light; see \cite{oren1995}.) To remove
shadows and specularities in face images, we can estimate this nine-dimensional
linear subspace and then project the images onto it. For \texttt{T-Rex}, we
estimate the subspace with the range of $F$, where $F$ is obtained from
Algorithm \ref{alg-Tyler-FA-EM} with $r = 9$. To benchmark our approach, we also
fit a nine-dimensional linear subspace using PCA, \emph{spherical PCA}
\cite{Locantore1999, Maronna2005} (\texttt{S-PCA}) which is a version of PCA
that aims to improve the robustness to outliers, \emph{outlier pursuit}
(\texttt{OP}) \cite{Xu2010, McCoy2011} which is based on convex optimization and
nuclear norm minimization, and \emph{S-Reaper} \cite{Lerman2015} which is
another convex optimization-based approach. We also tried to apply \texttt{STE}
\cite{yu2024subspace} to this problem using the code provided by the authors
(\url{https://github.com/alexfengg/STE}), but \texttt{STE} ran out of memory
while computing the spectral decomposition of a matrix of size $32256 \times
32256$.

\begin{figure}[!htb]
\centering
\subfloat{\includegraphics[width=0.67\textwidth]{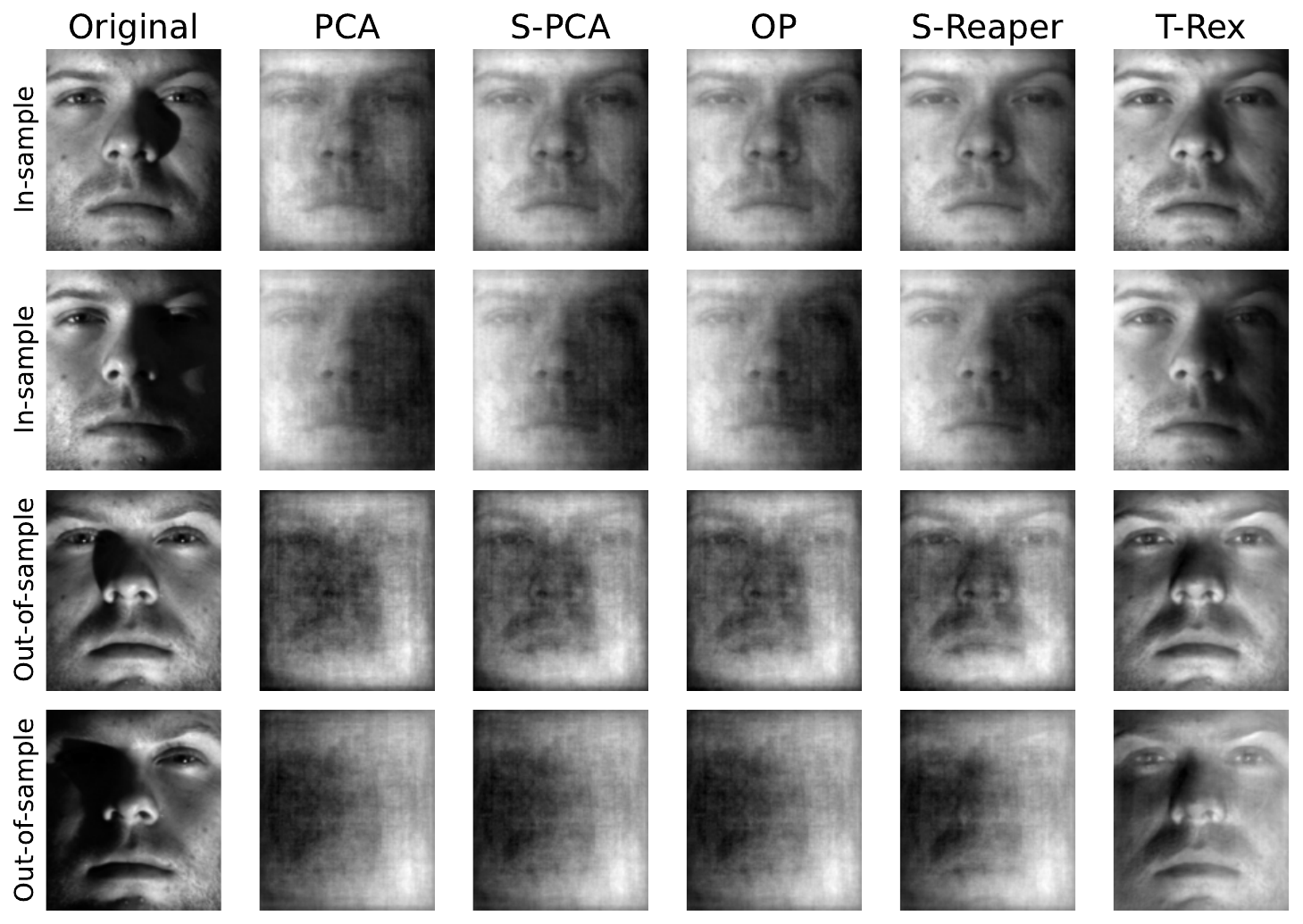}}
\caption{Face images projected onto nine-dimensional linear models. The
original images (leftmost column) are projected onto subspaces that were
fitted using five different methods. The first two rows show two in-sample
faces. The last two rows show two out-of-sample faces. (The out-of-sample
images were not used to fit the linear subspaces.)}
\label{fig:recovered-faces}
\end{figure}

\paragraph{Results.} 
Figure \ref{fig:recovered-faces} shows several images from the dataset
projected onto the computed nine-dimensional subspaces, with the centering added
back after the projection. We see that the linear subspace obtained from
\texttt{T-Rex} effectively removes shadows and produces the
clearest images, particularly for the out-of-sample images. We have included all
32 out-of-sample face images in Appendix \ref{sec:all-face-images}.

\begin{figure}[!htb]
\centering
\subfloat{\includegraphics[width=0.47\textwidth]{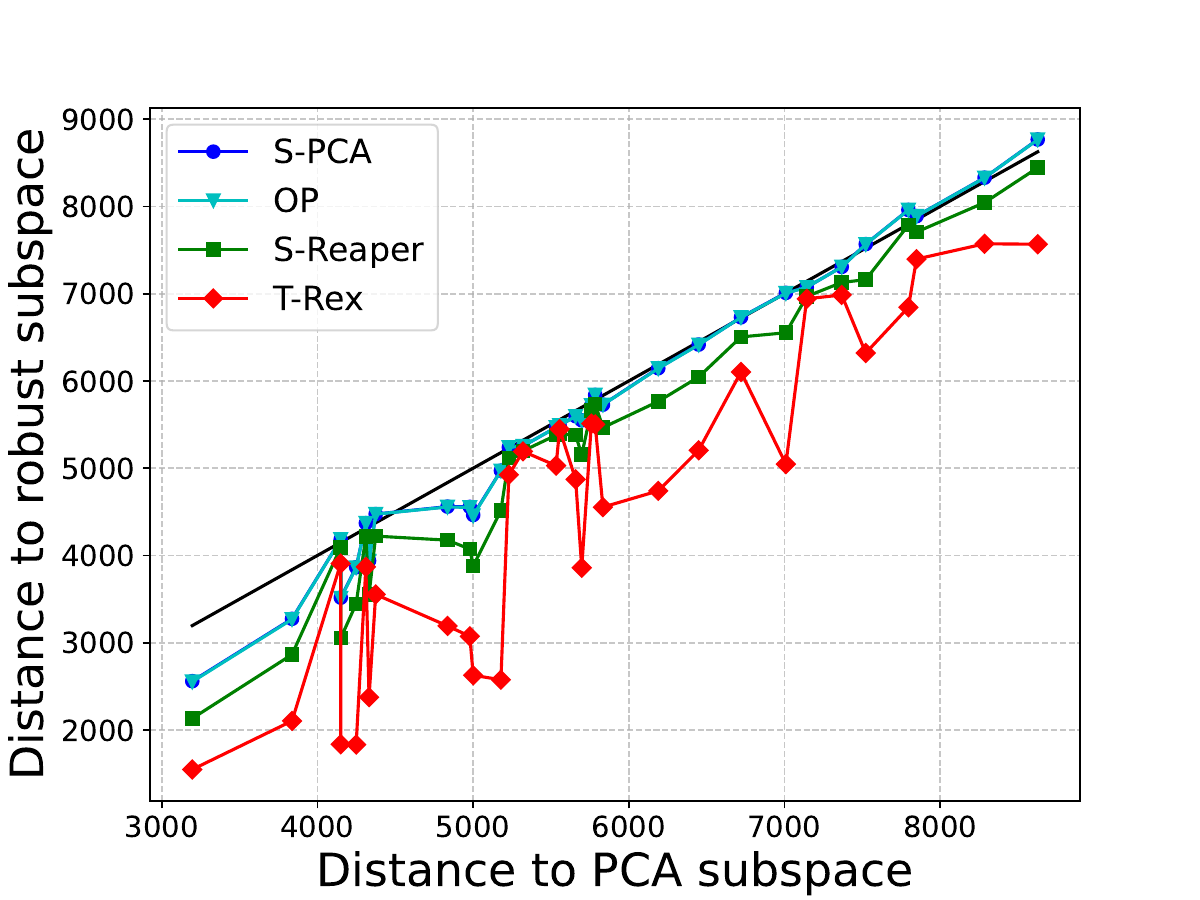}} 
\caption{The ordered distances of the out-of-sample face images to each robust
subspace as a function of the ordered distances to the PCA subspace. The
subspaces were computed from 32 in-sample images; this graph shows how the model
generalizes to the 32 out-of-sample images. Lower values are better. }
\label{fig:robust-distances}
\end{figure}

To quantitatively compare the different approaches, we show the distances of the
32 out-of-sample images to each of the fitted subspaces in Figure
\ref{fig:robust-distances}. Points below
the straight line indicate that the corresponding image is closer to the
subspace fitted by a robust method than to the subspace fitted using PCA.
Smaller values are better, and we see that
\texttt{T-Rex} outperforms the competitors, achieving the smallest
distance for 30 out of the 32 test images. 

\section{Conclusions}
\label{sec:conclusions}
We presented \texttt{T-Rex}, a scalable algorithm for solving Tyler's ML
estimation problem while enforcing a low-rank plus diagonal covariance
structure. In our experiments, \texttt{T-Rex} shows impressive and
robust performance, and we therefore believe that \texttt{T-Rex} contributes to
a growing set of tools for robust high-dimensional statistics. In the future, we
plan to extend \texttt{T-Rex} to support weighted ML estimation, and also study
the effect of regularization and shrinkage \cite{Chen2011, Sun2014, Ollila2021}.

\newpage
\appendix 

\section{Appendix}

\subsection{The latent variable model}
\label{sec:latent-variable-model-appendix}
Here we provide a heuristic motivation for the choice of joint density
\eqref{e:latent-variable-model} for the latent variable model. Suppose $Z \sim
N(0, \Sigma)$ is an $n$-dimensional random variable and let $\mathcal{A}_1
\subseteq \reals^n$. The pdf of $Z$, denoted by $p_Z(z)$, is defined through 
\[
\Prob(Z \in \mathcal{A}_1) = \int_{\mathcal{A}_1} p_Z(z) dz.
\] Consider the change of variables $z = rx$ where $r > 0$ and $x$ is restricted to the unit sphere $\mathbb{S}^{n-1}$. Let $\mathcal{A}_{2} \subseteq \mathbb{S}^{n-1} \times \reals_+$ be the set of all $(x, r)$ such that $z = rx$ for some $z \in \mathcal{A}_1$. We can express the volume element $dz$ as  $dz = r^{n-1} dx dr$. Hence,
\[
\Prob(Z \in \mathcal{A}_1) = \int_{\mathcal{A}_{2}} p_Z(rx) r^{n-1} dx dr.
\] This should be equal to $\Prob((X, R) \in \mathcal{A}_{2})$,
so the joint pdf of $(X, R)$ is $p_{X, R}(x, r) = p_Z(rx) r^{n-1}$, which is the
same as \eqref{e:latent-variable-model}.

\subsection{Implementation details}
\label{sec:Implementation details}
\paragraph{Initialization.}
We initialize \texttt{T-Rex} using the following
strategy based on PCA of the correlation matrix. (One reason for applying PCA to
the sample correlation matrix, rather than the sample covariance matrix itself,
is to ensure the initialization is scale-invariant.) Let $S \in \symm^n$ be the
sample covariance matrix, and define the vector of standard derivations as $s =
\diag(S)^{1/2}$. The sample correlation matrix is then given by $R =
\diag(s)^{-1} S \diag(s)^{-1}$. First we compute the spectral decomposition $R =
Q \Lambda Q^T$ at a cost of order $\mathcal{O}(n^3)$, and then form the low-rank
approximation $\hat{R} = \hat{F} \hat{F}^T + \hat{D}$, where 
\[
\hat{F} = Q_{1:r} \Lambda^{1/2}_{1:r}, \qquad \hat{D} = \diag(\diag(R - \hat{F} \hat{F}^T)).
\] The corresponding approximation of the sample covariance matrix is $\hat{S} = F_0 F_0^T + D_0$, where 
\[
F_0 = \diag(s) \hat{F}, \qquad D_0 = \diag(s) \hat{D} \diag(s).
\] We use this choice of $F_0$ and $D_0$ to initialize Algorithm \ref{alg-Tyler-FA-EM}.
This initialization is not specific for Algorithm \ref{alg-Tyler-FA-EM} and can
be used to initialize any iterative method for fitting a statistical factor
model. The same initialization strategy was used in \cite{Johansson23}, and in
statistics it is closely related to what is called 
\emph{principal component factor
analysis} \cite[\S9.3]{johnson2007applied}.

In the robust subspace recovery example in \S\ref{sec:num exp robust subspace
recovery} where $n \gg m$, we want to avoid forming an $n \times n$ matrix. We
therefore initialize \texttt{T-Rex} using PCA on the data matrix. In practice,
this can be done by computing the reduced SVD of the data matrix at
a cost of order $\mathcal{O}(nm^2)$.

\paragraph{Termination criterion.} A termination criterion that often works well
in practice is to terminate when the relative change in objective value between 
two consecutive iterates is smaller than a threshold. Specifically, 
we can terminate Algorithm \ref{alg-Tyler-FA-EM} whenever
\BEQ \label{e:relative-opt-gap-criterion}
\frac{|f(F_{k+1}, D_{k+1}) - f(F_k, D_k)|}{|f(F_k, D_k)|} \leq \epsilon
\EEQ for some prespecified $\epsilon > 0$, where $f(F, D) = \log\det(FF^T + D) +
(n/m)\sum_{i=1}^m \log(x_i^T(FF^T + D)^{-1} x_i)$ is the objective function of
the maximum likelihood estimation problem \eqref{e:Tyler-prob-factor-model}. 
To evaluate the first term of $f(F, D)$ efficiently, without incurring a cost that is cubic in $n$,
we use the identity
\[
\log \det (FF^T + D) = \log \det (D) + \log \det(I + F^T D^{-1} F).
\] The cost of forming the $r \times r$ matrix $I + F^T D^{-1} F$ is of order $\mathcal{O}(nr^2)$.
Evaluating $\log \det(I + F^T D^{-1} F)$ is cheap since 
the matrix $I + F^T D^{-1} F$ is of dimension $r$. (In practice, we evaluate the
log-determinant by computing the Cholesky factorization of $I + F^T D^{-1} F$.)

To compute the second term of $f(F, d)$, we
must evaluate $(FF^T + D)^{-1} x_i$ for $i = 1, \dots, m$. This can be done
at a cost of order $\mathcal{O}(nmr)$ using the matrix inversion formula \cite{boyd2004}.


\paragraph{EM for Gaussian factor analysis.}
The EM algorithm by Rubin \& Thayer \cite{Rubin1982} is perhaps the most well-known
method for solving the Gaussian ML estimation problem
\eqref{e:Gaussian-FA-prob}. Its iteration complexity of order $\mathcal{O}(n^2
r)$ sets it apart from many competing approaches. An additional advantage of the algorithm
lies in its simplicity, as it can be implemented with around 10 lines of
code in a high-level programming language. 

Rubin and Thayer's EM algorithm can be summarized as follows. Given an iterate
$(F_k, D_k)$, the next iterate is given by 
\[
\begin{split}
F_{k+1} & = S A_k B_k^{-1}, \\
D_{k+1} & = \diag(\diag(S - 2 S A_k F_{k+1}^T + F_{k+1} B_k F_{k+1}^T)), 
\end{split}
\] where 
\[
\begin{split}
A_k & = D_k^{-1} F_k H_k \\
B_k & = H_k + A_k^T S A_k \\
H_k & = (I + F_k^T D_k^{-1} F_k)^{-1}.
\end{split}
\]

\subsection{Experimental details}
\label{sec:exp-details-covariance-construction}
For the experiments in \S\ref{sec:num exp synthetic}, we assume that the true covariance matrix is of
the form $\Sigma_{\text{true}} = F_{\text{true}} F_{\text{true}}^T +
D_{\text{true}}$. To construct $\Sigma_{\text{true}}$ for a given value of
$n$, we first compute the sample covariance matrix $S \in
\symm^n$ from the daily returns of $n$ companies on S\&P 500 over the period
from January 1, 2022 to January 1, 2024. We then apply PCA on $S$ and choose the
columns of the factor loading matrix $F_{\text{true}}$ to be equal to the $r$
first principal directions, scaled with the square root of the corresponding
eigenvalues, and then pick $D_{\text{true}} = \diag(\diag(S - F_{\text{true}}
F_{\text{true}}^T))$. 

\subsection{Face images}
\label{sec:all-face-images}
\begin{figure*}[!htb]
\centering
\begin{minipage}{0.48\textwidth}
\centering
\includegraphics[width=\textwidth]{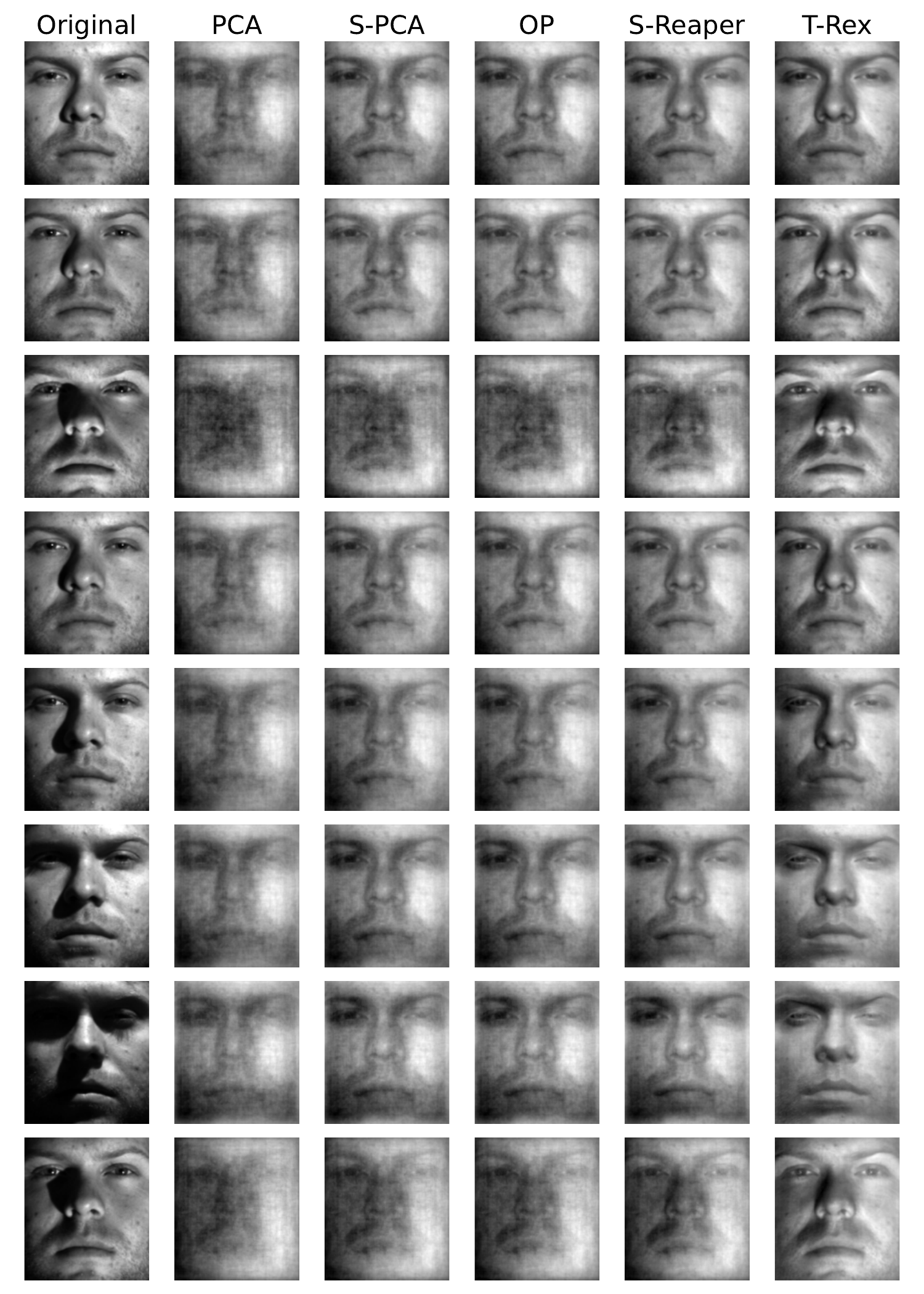}
\caption{Out-of-sample test images 1-8 projected onto nine-dimensional linear subspaces.}
\label{fig:test-image-1-appendix}
\end{minipage}\hfill
\begin{minipage}{0.48\textwidth}
\centering
\includegraphics[width=\textwidth]{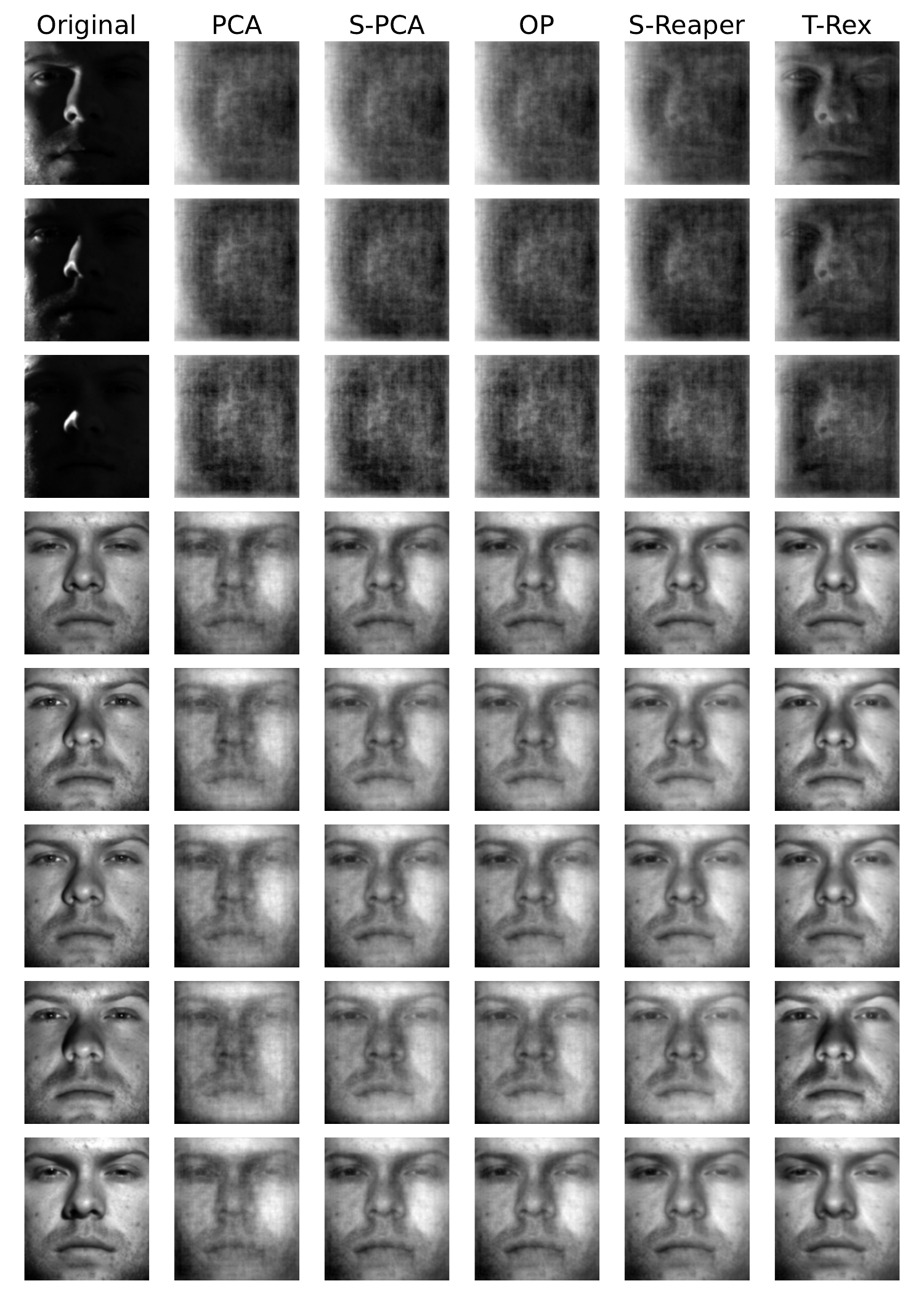}
\caption{Out-of-sample test images 9-16 projected onto nine-dimensional linear subspaces.}
\label{fig:test-image-2-appendix}
\end{minipage}
\end{figure*}

\begin{figure*}[!htb]
\centering
\begin{minipage}{0.48\textwidth}
\centering
\includegraphics[width=\textwidth]{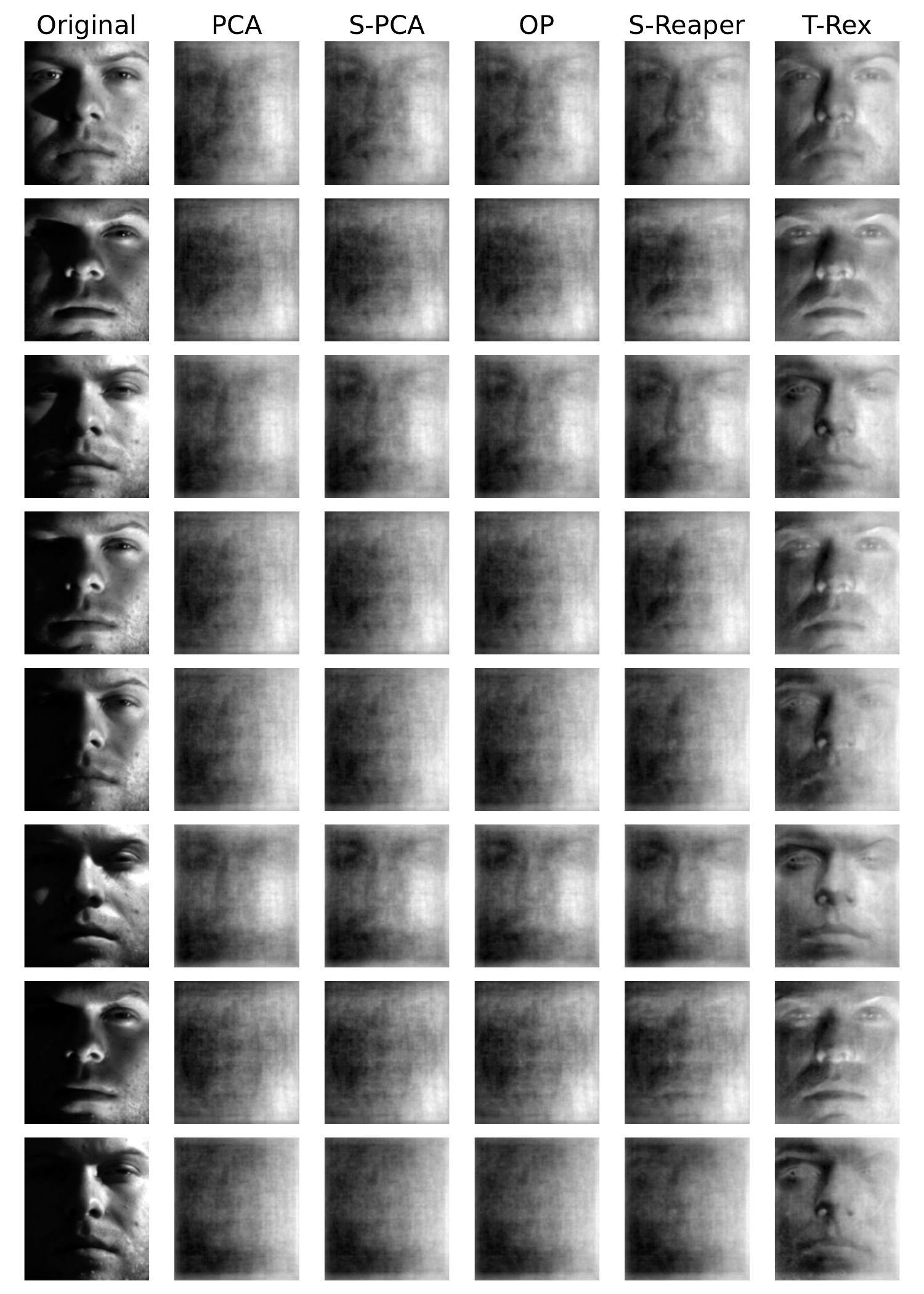}
\caption{Out-of-sample test images 17-24 projected onto nine-dimensional linear subspaces.}
\label{fig:test-image-3-appendix}
\end{minipage}\hfill
\begin{minipage}{0.48\textwidth}
\centering
\includegraphics[width=\textwidth]{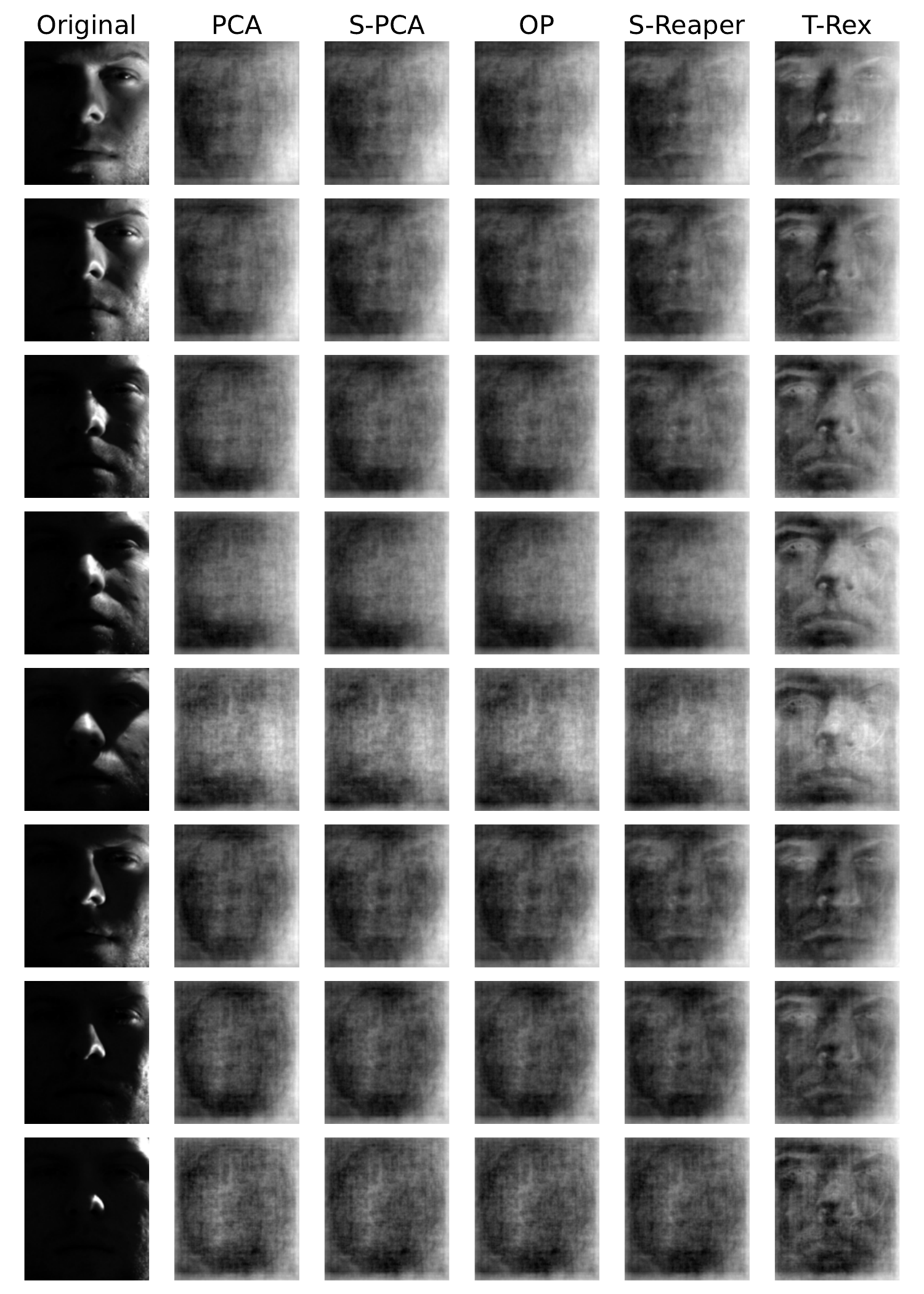}
\caption{Out-of-sample test images 25-32 projected onto nine-dimensional linear subspaces.}
\label{fig:test-image-4-appendix}
\end{minipage}
\end{figure*}

In Figures \ref{fig:test-image-1-appendix}-\ref{fig:test-image-4-appendix} we
show all original out-of-sample images together with their projections onto
nine-dimensional subspaces learned using five different modeling
techniques. (The subspaces were fitted on the training data.) We see that
\texttt{T-Rex} produces clearer images than the other methods, and often significantly so.

\newpage
\bibliographystyle{alpha}
\bibliography{references.bib}

\newcommand{\etalchar}[1]{$^{#1}$}
\begin{thebibliography}{HFBM{\etalchar{+}}23}

\bibitem[Bas94]{basilevsky1994}
A.~Basilevsky.
\newblock {\em {Statistical Factor Analysis and Related Methods: Theory and
  Applications}}.
\newblock John Wiley \& Sons, New York, 1994.

\bibitem[BBG{\etalchar{+}}21]{Bouchard2021}
F.~Bouchard, A.~Breloy, G.~Ginolhac, A.~Renaux, and F.~Pascal.
\newblock {A Riemannian Framework for Low-Rank Structured Elliptical Models}.
\newblock {\em IEEE Trans. Signal Process.}, 69:1185--1199, 2021.

\bibitem[BJ03]{Basri2003}
R.~Basri and D.W. Jacobs.
\newblock {Lambertian reflectance and linear subspaces}.
\newblock {\em IEEE Trans. Pattern Anal. Mach. Intell.}, 25(2):218--233, 2003.

\bibitem[BKM11]{Bartholomew2011}
D.~Bartholomew, M.~Knott, and I.~Moustaki.
\newblock {\em {Latent variable models and factor analysis: A unified
  approach}}.
\newblock John Wiley \& Sons, 2011.

\bibitem[BV04]{boyd2004}
Stephen Boyd and Lieven Vandenberghe.
\newblock {\em Convex optimization}.
\newblock Cambridge university press, 2004.

\bibitem[CBB{\etalchar{+}}21]{Collas2021}
A.~Collas, F.~Bouchard, A.~Breloy, G.~Ginolhac, C.~Ren, and J.~Ovarlez.
\newblock {Probabilistic PCA From Heteroscedastic Signals: Geometric Framework
  and Application to Clustering}.
\newblock {\em IEEE Trans. Signal Process.}, 69:6546--6560, 2021.

\bibitem[Chr07]{christian2007}
Walck Christian.
\newblock Handbook on statistical distributions for experimentalists.
\newblock {\em University of Stockholm}, 2007.

\bibitem[Con95]{connor1995}
G.~Connor.
\newblock The three types of factor models: A comparison of their explanatory
  power.
\newblock {\em Financial Analysts Journal}, 51(3):42--46, 1995.

\bibitem[CWH11]{Chen2011}
Yilun Chen, Ami Wiesel, and Alfred~O Hero.
\newblock Robust shrinkage estimation of high-dimensional covariance matrices.
\newblock {\em IEEE Trans. Signal Process.}, 59(9):4097--4107, 2011.

\bibitem[DG22]{Danon2022}
L.~Danon and D.~Garber.
\newblock {Frank-Wolfe-based Algorithms for Approximating Tyler's M-estimator}.
\newblock In S.~Koyejo, S.~Mohamed, A.~Agarwal, D.~Belgrave, K.~Cho, and A.~Oh,
  editors, {\em Adv. Neural Inf. Process. Syst}, volume~35, pages 3637--3648.
  Curran Associates, Inc., 2022.

\bibitem[DLR77]{dempster1977}
A.~Dempster, N.~Laird, and D.~Rubin.
\newblock {Maximum likelihood from incomplete data via the EM algorithm}.
\newblock {\em J. Roy. Stat. Soc. Ser. B (Methodol.)}, 39(1):1--22, 1977.

\bibitem[FFFP04]{FeiFei2004}
L.~Fei-Fei, R.~Fergus, and P.~Perona.
\newblock {Learning Generative Visual Models from Few Training Examples: An
  Incremental Bayesian Approach Tested on 101 Object Categories}.
\newblock In {\em Proc. 2004 Conf. Comput. Vis. Pattern Recognit. Workshop},
  2004.

\bibitem[FM20]{Frank2020}
W.~Franks and A.~Moitra.
\newblock {Rigorous Guarantees for Tyler’s M-Estimator via Quantum
  Expansion}.
\newblock In Jacob Abernethy and Shivani Agarwal, editors, {\em Proc. 33rd
  Conf. Learn. Theory (COLT)}, volume 125 of {\em Proceedings of Machine
  Learning Research}, pages 1601--1632. PMLR, 09--12 Jul 2020.

\bibitem[FWZZ21]{Fan2021}
Jianqing Fan, Kaizheng Wang, Yiqiao Zhong, and Ziwei Zhu.
\newblock {Robust high dimensional factor models with applications to
  statistical machine learning}.
\newblock {\em Statist. Sci.}, 36(2):303, 2021.

\bibitem[GBC16]{Goodfellow2016}
Ian Goodfellow, Yoshua Bengio, and Aaron Courville.
\newblock {\em Deep Learning}.
\newblock MIT Press, 2016.
\newblock \url{http://www.deeplearningbook.org}.

\bibitem[GG13]{Greco2013}
M.~Greco and F.~Gini.
\newblock {Cramér-Rao Lower Bounds on Covariance Matrix Estimation for Complex
  Elliptically Symmetric Distributions}.
\newblock {\em IEEE Trans. Signal Process.}, 61(24):6401--6409, 2013.

\bibitem[GLN20]{Goes2020}
John Goes, Gilad Lerman, and Boaz Nadler.
\newblock {Robust sparse covariance estimation by thresholding Tyler’s
  M-estimator}.
\newblock {\em Ann. Statist.}, 48(1):86 -- 110, 2020.

\bibitem[GR07]{gradshteyn2007}
Izrail~Solomonovich Gradshteyn and Iosif~Moiseevich Ryzhik.
\newblock {\em Table of integrals, series, and products}.
\newblock Academic press, 2007.

\bibitem[HFBM{\etalchar{+}}23]{hippert2023learning}
A.~Hippert-Ferrer, F.~Bouchard, A.~Mian, T.~Vayer, and A.~Breloy.
\newblock {Learning Graphical Factor Models with Riemannian Optimization}.
\newblock In {\em Proc. Joint Eur. Conf. Mach. Learn. Knowl. Discov.
  Databases}, pages 349--366. Springer, 2023.

\bibitem[HFEBG22]{Hippert2021}
A.~Hippert-Ferrer, M.~N. {El Korso}, A.~Breloy, and G.~Ginolhac.
\newblock {Robust low-rank covariance matrix estimation with a general pattern
  of missing values}.
\newblock {\em Signal Process.}, 195:108460, 2022.

\bibitem[HGBF21]{hong2021heppcat}
D.~Hong, K.~Gilman, L.~Balzano, and J.~Fessler.
\newblock {HePPCAT: Probabilistic PCA for data with heteroscedastic noise}.
\newblock {\em IEEE Trans. Signal Process.}, 69:4819--4834, 2021.

\bibitem[HJ66]{Harman1966}
H.~H. Harman and W.~H. Jones.
\newblock {Factor Analysis by Minimizing Residuals (Minres)}.
\newblock {\em Psychometrika}, 31(3):351--368, 1966.

\bibitem[JG72]{Joreskog1972}
K.~G. Jöreskog and A.~S. Goldberger.
\newblock Factor analysis by generalized least squares.
\newblock {\em Psychometrika}, 37(3):243--260, 1972.

\bibitem[JJ97]{jamshidian1997}
M.~Jamshidian and R.~Jennrich.
\newblock {Acceleration of the EM algorithm by using quasi-Newton methods}.
\newblock {\em J. Roy. Stat. Soc. Ser. B (Stat. Methodol.)}, 59(3):569--587,
  1997.

\bibitem[JOP{\etalchar{+}}23]{Johansson23}
K.~Johansson, M.~Ogut, M.~Pelger, T.~Schmelzer, and S.~Boyd.
\newblock {A Simple Method for Predicting Covariance Matrices of Financial
  Returns}.
\newblock {\em Found. Trends Econometrics}, 12(4):324--407, 2023.

\bibitem[JW07]{johnson2007applied}
R.A. Johnson and D.W. Wichern.
\newblock {\em {Applied Multivariate Statistical Analysis}}.
\newblock Pearson Prentice Hall, 2007.

\bibitem[Jö67]{Joreskog1967}
K.~Jöreskog.
\newblock {Some Contributions To Maximum Likelihood Factor Analysis}.
\newblock {\em Psychometrika}, 32(4):443--482, 1967.

\bibitem[KCHK05]{Lee2005}
L.~Kuang-Chih, J.~Ho, and D.J. Kriegman.
\newblock {Acquiring linear subspaces for face recognition under variable
  lighting}.
\newblock {\em IEEE Trans. Pattern Anal. Mach. Intell.}, 27(5):684--698, 2005.

\bibitem[KM19]{Khamaru2019}
K.~Khamaru and R.~Mazumder.
\newblock {Computation of the maximum likelihood estimator in low-rank factor
  analysis}.
\newblock {\em Math. Prog.}, 176(1):279--310, 2019.

\bibitem[LCHG16]{Liao2016}
B.~Liao, S.~Chan, L.~Huang, and C.~Guo.
\newblock {Iterative Methods for Subspace and DOA Estimation in Nonuniform
  Noise}.
\newblock {\em IEEE Trans. Signal Process.}, 64(12):3008--3020, 2016.

\bibitem[LM18]{Lerman2018}
G.~Lerman and T.~Maunu.
\newblock {An Overview of Robust Subspace Recovery}.
\newblock {\em Proc. IEEE}, 106(8):1380--1410, 2018.

\bibitem[LMS{\etalchar{+}}99]{Locantore1999}
N.~Locantore, J.~Marron, D.~Simpson, N.~Tripoli, J.~Zhang, and K.~Cohen.
\newblock {Robust Principal Component Analysis for Functional Data}.
\newblock {\em Test}, 8(1):1--73, 1999.

\bibitem[LMTZ15]{Lerman2015}
G.~Lerman, D.~McCoy, J.~Tropp, and T.~Zhang.
\newblock {Robust Computation of Linear Models by Convex Relaxation}.
\newblock {\em Found. Comput. Math.}, 15(2):363--410, 2015.

\bibitem[LR95]{Liu1995}
C.~Liu and D.~Rubin.
\newblock {ML Estimation of the $t$-Distribution Using EM and Its Extensions,
  ECM and ECME}.
\newblock {\em Statist. Sinica}, 5(1):19--39, 1995.

\bibitem[LR98]{liu1998a}
C.~Liu and D.~Rubin.
\newblock {Maximum likelihood estimation of factor analysis using the ECME
  algorithm with complete and incomplete data}.
\newblock {\em Statist. Sinica}, pages 729--747, 1998.

\bibitem[LRW98]{liu1998}
C.~Liu, D.~Rubin, and Y.~Wu.
\newblock {Parameter expansion to accelerate EM: the PX-EM algorithm}.
\newblock {\em Biometrika}, 85(4):755--770, 1998.

\bibitem[Mar05]{Maronna2005}
R.~Maronna.
\newblock {Principal Components and Orthogonal Regression Based on Robust
  Scales}.
\newblock {\em Technometrics}, 47(3):264--273, 2005.

\bibitem[MGO22]{Mecklenbrauker2022}
C.~F. Mecklenbräuker, P.~Gerstoft, and E.~Ollila.
\newblock {DOA M-Estimation Using Sparse Bayesian Learning}.
\newblock In {\em IEEE Int. Conf. on Acoust., Speech Signal Process.}, pages
  4933--4937, 2022.

\bibitem[MGOP24]{Mecklenbrauker2024}
C.~F. Mecklenbräuker, P.~Gerstoft, E.~Ollila, and Y.~Park.
\newblock {Robust and sparse M-estimation of DOA}.
\newblock {\em Signal Process.}, 220:109461, 2024.

\bibitem[MPFO13]{Mahot2013}
M.~Mahot, F.~Pascal, P.~Forster, and J.~P. Ovarlez.
\newblock {Asymptotic Properties of Robust Complex Covariance Matrix
  Estimates}.
\newblock {\em IEEE Trans. Signal Process.}, 61(13):3348--3356, 2013.

\bibitem[MT11]{McCoy2011}
Michael McCoy and Joel Tropp.
\newblock Two proposals for robust {PCA} using semidefinite programming.
\newblock {\em Electron. J. Stat.}, 5:1123--1160, 2011.

\bibitem[OK03]{Ollila2003}
E.~Ollila and V.~Koivunen.
\newblock {Robust antenna array processing using M-estimators of
  pseudo-covariance}.
\newblock In {\em 14th IEEE Proc. on Personal, Indoor and Mobile Radio
  Commun.}, volume~3, pages 2659--2663, 2003.

\bibitem[ON95]{oren1995}
Michael Oren and Shree~K Nayar.
\newblock {Generalization of the Lambertian model and implications for machine
  vision}.
\newblock {\em Int. J. Comput. Vis.}, 14:227--251, 1995.

\bibitem[OPP21]{Ollila2021}
Esa Ollila, D.~Palomar, and Frédéric Pascal.
\newblock {Shrinking the Eigenvalues of M-Estimators of Covariance Matrix}.
\newblock {\em IEEE Trans. Signal Process.}, 69:256--269, 2021.

\bibitem[PG01]{Pesavento2001}
M.~Pesavento and A.B. Gershman.
\newblock {Maximum-Likelihood Direction-of-Arrival Estimation in the Presence
  of Unknown Nonuniform Noise}.
\newblock {\em IEEE Trans. Signal Process.}, 49(7):1310--1324, 2001.

\bibitem[PRFC03]{PISON2003}
G.~Pison, P.~J. Rousseeuw, P.~Filzmoser, and C.~Croux.
\newblock {Robust factor analysis}.
\newblock {\em J. Multivariate Anal.}, 84(1):145--172, 2003.

\bibitem[PZW{\etalchar{+}}23]{Palomar23}
D.~P. Palomar, R.~Zhou, X.~Wang, F.~Pascal, and E.~Ollila.
\newblock {\em {fitHeavyTail: Mean and Covariance Matrix Estimation under Heavy
  Tails}}, 2023.
\newblock R package version 0.2.0.

\bibitem[RT82]{Rubin1982}
D.~Rubin and D.~Thayer.
\newblock {EM algorithms for ML factor analysis}.
\newblock {\em Psychometrika}, 47(1):69--76, March 1982.

\bibitem[SB23]{Stoica2023}
P.~Stoica and P.~Babu.
\newblock {Low-Rank Covariance Matrix Estimation for Factor Analysis in
  Anisotropic Noise: Application to Array Processing and Portfolio Selection}.
\newblock {\em IEEE Trans. Signal Process.}, 71:1699--1711, 2023.

\bibitem[SBP14]{Sun2014}
Y.~Sun, P.~Babu, and D.~Palomar.
\newblock {Regularized Tyler's Scatter Estimator: Existence, Uniqueness, and
  Algorithms}.
\newblock {\em IEEE Trans. Signal Process.}, 62(19):5143--5156, 2014.

\bibitem[SBP16]{Sun2015}
Y.~Sun, P.~Babu, and D.~Palomar.
\newblock {Robust Estimation of Structured Covariance Matrix for Heavy-Tailed
  Elliptical Distributions}.
\newblock {\em IEEE Trans. Signal Process.}, 64(14):3576--3590, 2016.

\bibitem[SM05]{stoica2005spectral}
P.~Stoica and R.L. Moses.
\newblock {\em {Spectral Analysis of Signals}}.
\newblock Pearson Prentice Hall, 2005.

\bibitem[STW16]{Soloveychik2016a}
I.~Soloveychik, D.~Trushin, and A.~Wiesel.
\newblock {Group Symmetric Robust Covariance Estimation}.
\newblock {\em IEEE Trans. Signal Process.}, 64(1):244--257, 2016.

\bibitem[SW14]{Soloveychik2014}
I.~Soloveychik and A.~Wiesel.
\newblock {Tyler's Covariance Matrix Estimator in Elliptical Models With Convex
  Structure}.
\newblock {\em IEEE Trans. Signal Process.}, 62(20):5251--5259, 2014.

\bibitem[TB99]{tipping1999}
M.~Tipping and C.~Bishop.
\newblock Probabilistic principal component analysis.
\newblock {\em J. R. Stat. Soc. Ser. B Methodol.}, 61(3):611--622, 1999.

\bibitem[Tyl87]{Tyler1987}
D.~Tyler.
\newblock {A Distribution-Free M-Estimator of Multivariate Scatter}.
\newblock {\em Ann. Statist.}, 15(1):234--251, 1987.

\bibitem[WZ15]{Wiesel2015}
A.~Wiesel and T.~Zhang.
\newblock {Structured Robust Covariance Estimation}.
\newblock {\em Found. Trends® Signal Process.}, 8(3):127--216, 2015.

\bibitem[XCS10]{Xu2010}
H.~Xu, C.~Caramanis, and S.~Sanghavi.
\newblock {Robust PCA via Outlier Pursuit}.
\newblock In J.~Lafferty, C.~Williams, J.~Shawe-Taylor, R.~Zemel, and
  A.~Culotta, editors, {\em Adv. Neural Inf. Process. Syst.}, volume~23. Curran
  Associates, Inc., 2010.

\bibitem[YZL24]{yu2024subspace}
Feng Yu, Teng Zhang, and Gilad Lerman.
\newblock A subspace-constrained tyler's estimator and its applications to
  structure from motion.
\newblock In {\em Proc. IEEE/CVF Conf. Comput. Vis. Pattern Recognit.}, pages
  14575--14584, 2024.

\bibitem[ZCS16]{Zhang2016}
Teng Zhang, Xiuyuan Cheng, and Amit Singer.
\newblock {Mar{\v{c}}enko--Pastur law for Tyler’s M-estimator}.
\newblock {\em J. Multivar. Anal.}, 149:114--123, 2016.

\bibitem[ZCW22]{Zhang2022}
A.~Zhang, T.~Cai, and Y~Wu.
\newblock {Heteroskedastic PCA: Algorithm, optimality, and applications}.
\newblock {\em Ann. Statist.}, 50(1):53 -- 80, 2022.

\bibitem[Zha15]{Zhang2015}
T.~Zhang.
\newblock {Robust subspace recovery by Tyler's M-estimator}.
\newblock {\em Inf. Inference: A J. IMA}, 5(1):1--21, 11 2015.

\bibitem[ZKCM12]{Zoubir2012}
A.~M. Zoubir, V.~Koivunen, Y.~Chakhchoukh, and M.~Muma.
\newblock {Robust Estimation in Signal Processing: A Tutorial-Style Treatment
  of Fundamental Concepts}.
\newblock {\em IEEE Signal Process. Mag.}, 29(4):61--80, 2012.

\bibitem[ZKOM18]{Zoubir2018}
A.~M. Zoubir, V.~Koivunen, E.~Ollila, and M.~Muma.
\newblock {\em {Robust Statistics for Signal Processing}}.
\newblock Cambridge University Press, 2018.

\bibitem[ZLKP20]{Zhou2019}
R.~Zhou, J.~Liu, S.~Kumar, and D.~Palomar.
\newblock {Robust Factor Analysis for Parameter Estimation}.
\newblock {\em Proc. Comput.-Aided Syst. Theory}, pages 3--11, 2020.

\bibitem[ZLL14]{Zhang2014}
J.~Zhang, J.~Li, and C.~Liu.
\newblock {Robust Factor Analysis Using the Multivariate $t$-Distribution}.
\newblock {\em Statist. Sinica}, 24, January 2014.

\bibitem[ZYJ08]{Zhao2008}
J.~Zhao, Philip~L. Yu, and Q.~Jiang.
\newblock {ML estimation for factor analysis: EM or non-EM?}
\newblock {\em Stat. Comput.}, 18:109--123, 2008.

\end{thebibliography}
\end{document}